\documentclass{article}

\PassOptionsToPackage{numbers, compress}{natbib}
\usepackage[preprint]{neurips24_style}

\usepackage{amsmath,amssymb,mathtools,bm,empheq,amsthm}

\theoremstyle{plain}
\newtheorem{theorem}{Theorem}[section]
\newtheorem{proposition}[theorem]{Proposition}
\newtheorem{lemma}[theorem]{Lemma}

\theoremstyle{definition}

\theoremstyle{remark}

\def\bal#1\eal{\begin{align}#1\end{align}} 

\def\abar{\bar{\alpha}} 
\newcommand{\etc}{\textit{etc}. }
\newcommand{\eg}{\textit{e}.\textit{g}.\ }

\newcommand{\br}[1]{\left[#1\right]} 
\newcommand{\pr}[1]{\left(#1\right)} 
\newcommand{\cbr}[1]{\left\{#1\right\}} 

\def\transp{\mathsf{T}} 

\def\mc{\mathcal}
\def\R{\mathbb{R}}

\def\ast{*}

\newcommand{\grad}[2]{\ensuremath{\nabla_{#2}#1}} 

\newcommand {\bbmtx}{\begin{bmatrix}} 
\newcommand {\ebmtx}{\end{bmatrix}} 

\DeclareMathOperator*{\diagonal}{diag} 
\newcommand{\diag}[1]{\diagonal\pr{#1}}
\DeclareMathOperator*{\trace}{tr} 
\newcommand{\tr}[1]{\trace\pr{#1}}

\DeclareMathOperator{\E}{\mathbb{E}}

\usepackage[T1]{fontenc}
\usepackage[utf8]{inputenc}
\usepackage{babel}
\usepackage[font=small]{caption}
\usepackage{multicol}
\usepackage{multirow}
\usepackage{colortbl}
\usepackage[utf8]{inputenc} 
\usepackage[T1]{fontenc}    
\usepackage{hyperref}       
\usepackage{url}            
\usepackage{booktabs}       
\usepackage{amsfonts}       
\usepackage{nicefrac}       
\usepackage{microtype}      
\usepackage{xcolor}         
\newcommand\tts\footnotesize
\newcommand\cts\footnotesize
\definecolor{Myblue}{rgb}{0.86,0.84,0.92}  
\definecolor{Mypink}{rgb}{0.98,0.84,0.85}  

\newcommand{\glb}[1]{\colorbox{Mypink}{#1}}
\newcommand{\lcl}[1]{\colorbox{Myblue}{#1}}
\newcommand{\floormod}[1]{\lfloor #1 \rfloor}

\usepackage{graphicx}
\usepackage{subfigure}
\usepackage{cleveref}
\usepackage{booktabs} 

\title{A Modular Conditional Diffusion Framework for Image Reconstruction}

\author{
  Magauiya Zhussip$^{\ast \ddag}$ \\
  MTS AI \\
  \And
  Iaroslav Koshelev$^{\ast \ddag}$ \\
  AI Foundation and Algorithm Lab \\
  \And
  Stamatis Lefkimmiatis$^{\ddag}$ \\
  MTS AI \\
}
\begin{document}

\maketitle
\def\thefootnote{$\ast$}\footnotetext{Equal contribution}
\def\thefootnote{$\ddag$}\footnotetext{Work performed while at AI Foundation and Algorithm Lab}

\vspace{-3mm}
\begin{abstract}
Diffusion Probabilistic Models (DPMs) have been recently utilized to deal with various blind image restoration (IR) tasks, where they have demonstrated outstanding performance in terms of perceptual quality. However, the task-specific nature of existing solutions and the excessive computational costs related to their training, make such models impractical and challenging to use for different IR tasks than those that were initially trained for. This hinders their wider adoption, especially by those who lack access to powerful computational resources and vast amount of training data. In this work we aim to address the above issues and enable the successful adoption of DPMs in practical IR-related applications. Towards this goal, we propose a modular diffusion probabilistic IR framework (DP-IR), which allows us to combine the performance benefits of existing pre-trained state-of-the-art IR networks and generative DPMs, while it requires only the additional training of a relatively small module (0.7M params) related to the particular IR task of interest. Moreover, the architecture of the proposed framework allows for a sampling strategy that leads to at least four times reduction of neural function evaluations without suffering any performance loss, while it can also be combined with existing acceleration techniques such as DDIM. We evaluate our model on four benchmarks for the tasks of burst JDD-SR, dynamic scene deblurring, and super-resolution. Our method outperforms existing approaches in terms of perceptual quality while it retains a competitive performance with respect to fidelity metrics.
\end{abstract}
\section{Introduction}
\label{sec:intro}
With the advent of deep learning we have witnessed outstanding results in a wide range of computer vision tasks~\cite{Zou2023object}, including many challenging blind image restoration (IR) problems~\cite{Zhai2023comprehensive} such as burst imaging~\cite{li2023ntire}, super-resolution (SR)~\cite{Chen2022single}, deconvolution~\cite{Ren2020neural}, \etc The standard approach for supervised learning in a blind IR setting involves training a feed-forward network that should estimate the latent image based on the available low-quality measurements. Such models are usually trained to maximize \textit{fidelity} metrics like PSNR or SSIM, but the visual quality of the resulting images is sub-optimal~\cite{blau2018perception}.
The inclusion of perceptual losses~\cite{johnson2016perceptual} to the objective can improve the visual results, but fails to convincingly address the problem. 

\begin{figure}[!t]
\vspace{-2mm}
    \centering
    \includegraphics[width=1\linewidth]{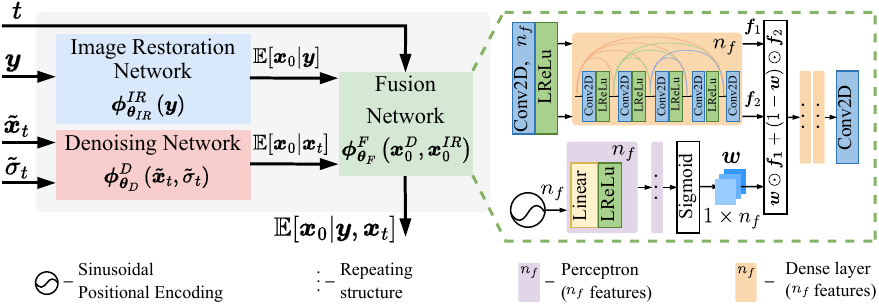}
    \caption{The proposed architecture consists of three modules: a Denoising Network $\bm \phi^D_{\bm \theta_D}\pr{\tilde{\bm x}_t, \tilde{\sigma}_t}$, an IR Network $\bm \phi^{IR}_{\bm \theta_{IR}}\pr{\bm y}$ and a Fusion Network $\bm \phi^F_{\bm \theta_F}\pr{\bm x_0^{IR}, \bm x_0^{D}, t}$. A small version of MIRNet~\cite{zamir2020learning} is used as the Denoising Network, while a pre-trained SwinIR~\cite{liang2021swinir} or BSRT~\cite{luo2022bsrt} or FFTFormer~\cite{kong2023efficient} is used as the IR Network, depending on the IR task.
    See~\cref{sec:network_architecture} for a detailed description.}
    \label{fig:denoiser_net}
\end{figure}

A promising direction towards IR results of high visual quality is to consider such problems within a generative framework. Several generative models have been recently proposed including Variational Autoencoders (VAEs)~\cite{kingma2013auto}, Generative Adversarial Neural Networks (GANs)~\cite{goodfellow2014generative}, Normalizing Flows (NFs)~\cite{dinh2014nice} and Diffusion Probabilistic Models (DPMs)~\cite{sohl-dickstein2015deep}. 
Due to their impressive results in image generation, they have been further utilized to perform \textit{conditional sampling} of high-quality images, with their low-quality or distorted counterparts playing the role of the conditional input~\cite{gatopoulos2020superresolution,ledig2017photo,lugmayr2020srflow,li2021srdiff}. To date, DPMs appear to be the most promising framework and lead to the best results among all existing generative approaches.

Nevertheless, there are certain limitations that existing DPMs face, which hinder their wider adoption in IR tasks. In particular, inference of such models involves a sampling process that requires a large number (in the order of hundreds) of neural function evaluations (NFEs), which can be computationally very expensive, especially when considering images of high resolution. Another important limitation is that an efficient conditioning on the image measurements has yet to be proposed for DPMs in order to make them applicable to a wider range of blind IR problems. Indeed, all of the existing methods aim to learn the parameters of a single input-conditioned network for a specific blind IR task. As a result, the trained model overfits on the distribution of the condition space, and the whole model has to be retrained if we need to employ it to a different reconstruction task than the one that was initially trained for. Considering the huge amount of data and computational resources required for training a single DPM (see~\cref{sec:appendix_dpm_requirements}), such re-training becomes infeasible if at least one of the previous requirements is not satisfied.\vspace{-1mm}

In this work we aim to address the above issues by proposing a novel conditional diffusion network coupled with an accelerated sampling process. Specifically, our network adopts an improved conditioning strategy and is built on the foundation of existing off-the-shelf IR networks paired with a denoising module, which is applicable to a variety of reconstruction problems without requiring any re-training. Additionally, we introduce an accelerated sampling procedure that is enabled by our proposed network architecture and allows the merging of a large number of sampling steps in a single one, computed with a single NFE. Our proposed acceleration can work in tandem with accelerated sampling schemes such as DDIM~\cite{song2021denoising}. To assess the performance of our network, we validate it on three challenging blind IR tasks, namely, burst joint demosaicking, denoising and super-resolution (JDD-SR), dynamic scene deblurring, and $4\times$ single image super-resolution (SISR). In all of the tested scenarios, our approach demonstrates the best perception-distortion trade-off among the state-of-the-art (SOTA) methods, while compared to other DPM-based solutions it requires a smaller number of sampling steps. 
\vspace{-2mm}
\section{Related Works}
\label{sec:related_works}
\paragraph{Burst Image Restoration.} \vspace{-2mm}
One of the pioneering works in multi-frame IR was introduced in~\cite{tsai1984multiframe}, where a frequency-domain-based solution was proposed. Then, several MAP models with various regularization terms have been designed to cope with visual artifacts caused by operating in the frequency domain~\cite{bascle1996motion, elad1997restoration, schultz1996extraction}. Using the same MAP framework, a JDD-SR method robust to noise and outliers was developed in~\cite{farsiu2004fast}. Meanwhile, the block matching alignment algorithm of~\cite{hasinoff2016burst} was extended by~\cite{wronski2019handheld} to obtain a robust motion model with the aid of an adaptive kernel interpolation method merging sparse pixel samples.\vspace{-1mm}

Advancements in deep learning have led to high-performing methods such as those in~\cite{dudhane2022burst, lecouat2021lucas, dbsr21, deeprep21, luo2021ebsr, lian2023kernel}. The DBSR approach~\cite{dbsr21} aligns multiple input frames in the feature space utilizing an optical flow estimator (\eg PWCNet~\cite{sun2018pwc}) and employs an attention-based fusion mechanism to aggregate features. In~\cite{lecouat2021lucas} a differentiable image registration module has been introduced, which exploits the aliasing effects appearing in bursts of low-resolution (LR) images. In~\cite{lian2023kernel} KBNet estimates blur kernels for a burst sequence to incorporate them with LR features so as to generate a better super-resolved image, while in~\cite{dudhane2022burst} BIPNet attempts to fuse complementary information from the burst sequence with the help of generated pseudo-burst features. Another line of work effectively employs deformable convolutions for the inter-frame alignment task~\cite{luo2021ebsr, wang2019edvr, dudhane2022burst} and achieves SOTA results in various tasks, including burst SR. 

\vspace{-1mm}
\paragraph{Single Image Restoration.}
Among the recent IR methods, the most successful ones are those that adopt an end-to-end supervised formulation, where a deep neural network is trained to directly map a low-quality and degraded image to a point estimate of the latent high-quality image~\cite{zhao2016loss, lim2017enhanced, tsai2022stripformer, zamir2022restormer}. Consequently, in the pursuit of further improving the reconstructions and achieving a better pixel-level result, more advanced network architectures have been proposed~\cite{zamir2021multi, liang2021swinir, chen2023activating, kong2023efficient}, at the cost of being more computationally heavy. While this formulation leads to SOTA fidelity (\eg, PSNR, SSIM), the produced output is an average/median of all plausible predictions, which typically lacks high-frequency information (\eg texture). 

Generative adversarial networks (GAN)~\cite{goodfellow2014generative} have been adopted by several IR methods such as SISR~\cite{ledig2017photo,wang2018esrgan,wang2021real} and dynamic scene deblurring ~\cite{kupyn2018deblurgan,kupyn2019deblurgan, zhang2020deblurring} to produce more natural and perceptually pleasing results. Although this adversarial non-reference formulation aims to push the predictions towards the manifold of natural images, it is also prone to introducing unrealistic texture and hallucinations in the output~\cite{cohen2018distribution}. Moreover, the adversarial training process requires extra supervision as it can easily fall into a mode collapse or may diverge~\cite{arora2017generalization, salimans2016improved}.

Likelihood-based deep generative models such as NFs~\cite{lugmayr2021ntire, lugmayr2020srflow}, auto-regressive models~\cite{guo2022lar}, and VAEs~\cite{prakash2020fully} have also been applied to IR tasks, where one can obtain a diverse set of predictions from a learned posterior~\cite{prakash2020fully}. Conditioned on LR inputs, flow-based methods attempt to map high-resolution (HR) images to the latent flow-space. Although such techniques circumvent the training instability met in GANs, strong architectural constraints (\eg network invertibility) still remain an issue. 

Recently another class of methods based on a stochastic diffusion process has been introduced and demonstrated outstanding performance on various tasks that range from unconditional image generation~\cite{ho2020denoising, nichol2021improved, rombach2022high} to image-to-image translation/restoration~\cite{li2021srdiff, saharia2021image, whang2022deblurring, gao2023implicit, rombach2022high, delbracio2023inversion, wang2023zeroshot, ren2023multiscale}. DvSR proposed in~\cite{whang2022deblurring} employs a “predict-and-refine” conditional diffusion method specifically tailored for the image deblurring task, while SRDiff~\cite{li2021srdiff} utilizes features of a pretrained SR model for conditional super-resolved image generation. Further, recent works in~\cite{wang2023zeroshot, Xia_2023_ICCV} have considered several IR tasks (\eg inpainting, super-resolution, colorization, etc.). In conclusion, their ability to capture complex statistics of the visual world, makes DPMs a very attractive solution that is worth being further investigated.

\vspace{-1mm}
\section{Proposed Conditional Diffusion Model}
\label{sec:proposed_method}
\vspace{-1mm}
\subsection{Background} \label{sec:diffusion_theory}
\vspace{-1mm}
Denoising Diffusion Probabilistic Models (DDPMs)~\cite{ho2020denoising,sohl-dickstein2015deep} are special cases of Hierarchical Markovian Variational Autoencoders where the dimension of the latent variables matches the dimension of the data. Starting with a sample $\bm x_0 \!\in\! \R^N$, the encoding sequence $\cbr{\bm x_t}_{t=0}^T$ traverses the latent space with a \textit{diffusion process} defined by a Gaussian transition probability: 
\bal\label{eq:diffusion_kernel}
q\pr{\bm x_t | \bm x_{t-1}} \equiv \mc N \pr{\bm x_t; \sqrt{1 - \beta_t} \bm x_{t-1}, \beta_t \bm I_N}.
\eal
The sequence $0 < \beta_1, \beta_2,\ldots,\beta_T < 1$ that appears in \cref{eq:diffusion_kernel} defines the noise scheduling for the forward process in such a way so that the latent variable at the final timestep $T$ approximates the standard Gaussian: $\bm x_T \sim \mc N\pr{\bm x_T; \bm 0, \bm I_N}$. Based on this diffusion process, it is possible to express the transition probability directly from $\bm x_0$ to $\bm x_t$ in closed form as:
\bal\label{eq:cumulative_forward}
q\pr{\bm x_t | \bm x_0} = \mc N \pr{\bm x_t; \sqrt{\abar_t} \bm x_0, \pr{1 - \abar_t} \bm I_N},
\eal
where $\alpha_t \equiv 1 - \beta_t$ and $\abar_t \equiv \prod_{s=1}^t \alpha_ s$. 

The \textit{reverse process} is enabled by the posterior distribution which is represented in the form: 
\bal \label{eq:posterior}
p \pr{\bm x_{t-1} | \bm x_t, \bm x_0} =  \mc N \pr{\bm x_{t-1}; \bm \mu_t\pr{\bm x_t, \bm x_0}, \sigma^2_t \bm I_N},
\eal
where $\bm \mu_t\pr{\bm x_t, \bm x_0} \equiv \frac{\sqrt{\abar_{t-1}}\beta_t}{1 - \abar_t} \bm x_0 + \frac{\sqrt{\alpha_t}\pr{1 - \abar_{t-1}}}{1 - \abar_t} \bm x_t$ and $\sigma^2_t \equiv \frac{1 - \abar_{t-1}}{1 - \abar_t} \beta_t$. DDPMs aim to approximate its mean by the quantity $\bm \mu_{\bm \theta} \pr{\bm x_t, t}$, which is learned from training data, and then utilize~\cref{eq:posterior} to perform sampling. There are different possible parameterizations of $\bm \mu_t\pr{\bm x_t, \bm x_0}$, which accordingly lead to different interpretations for the transition mean~\cite{luo2022understanding}. In this work, we pursue the one based on the score function $\nabla \log p\pr{\bm x_t}$, which reads as:
\bal\label{eq:mean_via_score}
\bm \mu_t\pr{\bm x_t, \bm x_0} = \frac{\bm x_t + \pr{1 - \alpha_t}\nabla \log{p\pr{\bm x_t}}}{\sqrt{\alpha_t}}.
\eal
In this case, the reverse process defined in~\cref{eq:posterior} can be considered as sampling via Annealed Langevin Dynamics, in which the score function is approximated by the quantity $\bm s_{\bm \theta} \pr{\bm x_t, t}$ learned via denoising score matching~\cite{hyvarinen2005estimation,vincent2011connection}.
\vspace{-2mm}
\subsection{Conditional Score Matching}\label{sec:conditional_score}
\vspace{-2mm}
The diffusion models described above do not take into account the dependence of the sampled data on their degraded observations $\bm y\!\in\!\R^M$, when we are dealing with IR problems. Fortunately, the score-based models can be extended to accommodate conditional sampling by replacing the score function in~\cref{eq:mean_via_score} with a conditional score function $\nabla_{\bm x_t} \log p\pr{\bm x_t | \bm y}$. For non-blind IR problems, a popular approach is to decompose the conditional score function into a score function $\nabla \log p\pr{\bm x_t}$ and a log-likelihood gradient term $\nabla_{\bm x_t} \log p\pr{\bm y | \bm x_t}$~\cite{nguyen2017plug,song2021scorebased}. This last term is directly dependent on the image formation model, which unfortunately is unknown for blind IR tasks. Therefore, most of the existing works~\cite{li2021srdiff,ren2023multiscale,rombach2022high} aim instead to learn the primal conditional score function $\nabla_{\bm x_t} \log p\pr{\bm x_t | \bm y}$ via \textit{ad-hoc} conditional denoising score matching.
In this work, we also utilize the primal conditional score function, but we rely on its explicit form as given in the following lemma, whose proof is provided in the~\cref{sec:score_function_proof}.
\begin{lemma}\label{lem:score_function}
Let $\bm y\!\in\!\R^{M}$, $\bm x_0\!\in\!\R^{N} \sim p\pr{\bm x_0|\bm y}$, and $\bm x_t\!\in\!\R^{N}$, $\abar_t\!\in\!\R$ are defined as in~\cref{eq:cumulative_forward}. Then, the conditional score function is computed as:
\bal\label{eq:score_via_expectation}
\vspace{-2mm}
\nabla_{\bm x_t} \log p\pr{\bm x_t | \bm y} = \frac{\sqrt{\abar_t} \E\br{\bm x_0 | \bm y, \bm x_t} - \bm x_t}{1 - \abar_t}.
\eal
\vspace{-6.8mm}
\end{lemma}
The above result implies that the conditional score function can be approximated by utilizing a trained joint reconstruction and denoising model. Specifically, 
if the augmented variable $\bm z_t = \bbmtx \bm y^\transp& \bm x_t^\transp\ebmtx^\transp$ represents the union of the degraded data $\bm y$ and the noisy data $\bm x_t$, then the conditional expected value $\E\br{\bm x_0 | \bm z_t}$ corresponds to the reconstructed underlying image $\bm x_0$ from the measurements $\bm z_t$.
 
A joint reconstruction model $\bm \phi_{\bm_\theta}\pr{\bm y, \bm x_t, t}$ can be trained by the minimization of the empirical expected pixel mean-squared error (MSE) across the samples from the training dataset $\mc D\pr{\bm x_0, \bm y, \bm x_t, t}$:
\vspace{-2mm}
\begin{align}
\label{eq:mmse_learning}
& \min_{\bm\theta} \E_{\bm x_0, \bm x_t, \bm y \sim \mc D} \lVert \bm \phi_{\bm \theta}\pr{\bm y, \bm x_t, t} - \bm x_0\rVert^2_2 = \min_{\bm\theta} \sum_i \lVert \bm \phi_{\bm \theta}\pr{\bm y^i, \bm x^i_t, t} - \bm x^i_0\rVert^2_2.
\end{align}
The optimal solution is the conditional expectation $\bm \phi^{\textrm{MSE}}_{\bm_\theta}\pr{\bm y, \bm x_t, t} = \E\br{\bm x_0 | \bm y, \bm x_t}$, and, thus, such a trained model can be substituted in~\cref{eq:score_via_expectation}. This amounts to approximating the conditional score function $\nabla_{\bm x_t} \log p\pr{\bm x_t | \bm y}$ with $\bm s_{\bm \theta}^c \pr{\bm y, \bm x_t, t} \equiv \frac{\sqrt{\abar_t} \bm \phi_{\bm_\theta}\pr{\bm y, \bm x_t, t} - \bm x_t}{1 - \abar_t}$.
\subsection{Proposed Network Architecture}\label{sec:network_architecture}
\vspace{-2mm}
Our objective is to parameterize the function $\bm \phi_{\bm_\theta}\pr{\bm y, \bm x_t, t}$ in a form of a neural network (CNN) and design a specific architecture of this network. The absence of explicit knowledge about the formation model $\bm x_0 \rightarrow \bm y$ requires the network to learn it implicitly from training data. Such an approach generally results in over-fitting, meaning that the trained model can only be employed for the task it was originally trained for~\cite{li2021srdiff,whang2022deblurring}. 

To overcome this problem, we initially build on the hypothesis that the conditional expectation $\E\br{\bm x_0 | \bm y, \bm x_t}$ related to the conditional score function in \cref{eq:score_via_expectation}, can be approximated by a function of two easier to compute conditional expectations, that is 
\bal
\E\br{\bm x_0 | \bm y, \bm x_t}\approx f\pr{\E\br{\bm x_0 | \bm y}, \E\br{\bm x_0 | \bm x_t}}.
\eal

Based on such an approximation, it is now possible to learn a single unconditional generative denoising model that can be applied in different reconstruction problems.
Further, we note that despite the absence of a good approximation of the likelihood term, $\E \br{\bm x_0 | \bm y }$, various task-specific networks trained with a fidelity objective are readily available in the literature. Indeed, using a similar reasoning as the one provided for~\cref{eq:mmse_learning}, such reconstruction networks can output a good approximation of the quantity $\E \br{\bm x_0 | \bm y }$. This finally motivates us to express the joint reconstruction and denoising network $\bm \phi_{\bm_\theta}\pr{\bm y, \bm x_t, t}$ into three components (see Figure~\ref{fig:denoiser_net}). Specifically, our network can be described as $\bm \phi^F_{\bm \theta_F}\pr{\bm \phi^{IR}_{\bm \theta_{IR}}\pr{\bm y}, \bm \phi^D_{\bm \theta_D}\pr{\tilde{\bm x}_t, \tilde{\sigma}_t}, t}$, where $\tilde{\bm x}_t \equiv \tfrac{\bm x_t}{\sqrt{\abar_t}}\sim \mc N\pr{\bm x_0, \tilde{\sigma}^2_t\bm I_N}$ is the noisy version of $\bm x_0$ with noise variance $\tilde{\sigma}^2_t \equiv \tfrac{1 - \abar_t}{\abar_t}$ according to~\cref{eq:cumulative_forward}, and the sub-modules $\bm \phi^D_{\bm \theta_D}, \bm \phi^{IR}_{\bm \theta_{IR}}, \bm \phi^F_{\bm \theta_F}$ are defined next.

\vspace{-1mm}
\noindent\textbf{IR network} $\bm \phi^{IR}_{\bm \theta_{IR}}\pr{\bm y}$, which is learned in a supervised manner to predict $\E\br{\bm x_0 | \bm y}$. Specifically, we employ the BSRT-Small~\cite{luo2022bsrt} for burst JDD-SR, FFTformer~\cite{kong2023efficient} for dynamic scene deblurring, and SwinIR~\cite{liang2021swinir} for SISR. We do not train these networks but use their publicly available trained weights.

\vspace{-1mm}
\noindent\textbf{Denoising network} $\bm \phi^D_{\bm \theta_D}\pr{\tilde{\bm x}_t, \tilde{\sigma}_t}$, which is learned in a supervised manner to predict $\E\br{\bm x_0 | \bm x_t}$ by denoising $\tilde{\bm x}_t$. Specifically, we employ a smaller version of MIRNet~\cite{zamir2020learning}, which we call MIRNet-S. It is obtained by reducing the amount of RRG and MRB blocks from the original architecture to three and one, respectively. We refer to the original paper~\cite{zamir2020learning} for the detailed description of the blocks structure, as we use them without any modifications. Once trained, this network is reused for all considered reconstruction problems. We note that we our motivation for utilizing a smaller version of MIRNet as a Denoising module, is to approximately match the number of parameters and the computational complexity of the networks used in our framework with those of the alternative methods under study. This way we can ensure a fair evaluation and comparison among competing methods. Such strategy has allowed us to achieve direct performance comparisons under similar conditions.

\vspace{-1mm}
\noindent\textbf{Fusion network} $\bm \phi^F_{\bm \theta_F}\pr{\bm x_0^{IR}, \bm x_0^{D}, t}$, which predicts the conditional expectation $\E\br{\bm x_0 | \bm y, \bm x_t}$. This module refines and combines the predictions of the previous two networks and is the only one that needs to be trained for each specific IR task. The fusion network accepts as inputs the image estimates $\bm x_0^{IR}, \bm x_0^{D}$ and a timestep $t$. Its architecture consists of two branches. The first one involves a convolution layer with $n_f$ output channels followed by a single dense block~\cite{huang2017densely} without batch normalization. Its purpose is to independently encode both input images into the corresponding features $\bm f_1, \bm f_2$ with $n_f$ channels each. The second branch encodes the timestep $t$ into a vector of weights $\bm w \in \pr{0, 1}^{n_f}$ using the sinusoidal positional encoding~\cite{vaswani2017attention}, followed by a two layer perceptron and a sigmoid function as the final activation. The features $\bm f_1, \bm f_2$ and the weights $\bm w$ are then passed to the Convex Combination Channel Attention (3CA) layer, which performs the per-channel aggregation of input features as a convex combination of the form: $\bm w \odot \bm f_1 + \pr{\bm 1 - \bm w} \odot \bm f_2$. The output of this layer is decoded by two consequent dense blocks with $n_f$ channels each, followed by a convolution layer which produces the final output $\E\br{\bm x_0 | \bm y, \bm x_t}$. This proposed architecture results in a significantly smaller network size than those of the Denoising and IR modules. Thus we can train the fusion network fast and by using only a small amount of problem-specific training data. While we explored several basic fusion architectures, we did not delve into extensive research to ascertain the optimal design. Our proposed fusion module serves as a proof of concept, validating our framework and demonstrating its potential for performance enhancement. A comprehensive investigation into optimal fusion architectures remains a promising area for future research. \vspace{-1mm}

Such a modular overall architecture allows us to capitalize on the existing SOTA non-blind denoising and blind IR networks, while it also allows us to easily replace any of these networks when better ones become available in the future. As we describe next, another important advantage of our proposed pipeline, is that it allows us to achieve a significant acceleration for the sampling process without incurring any loss of reconstruction quality. 

\vspace{-2mm}
\subsection{Proposed Accelerated Sampling}\label{sec:accelerated_sampling}
\vspace{-2mm}
According to~\cref{eq:posterior}, our conditional denoiser should be evaluated for all timesteps $t = \overline{T, ..., 0}$, which leads to a total of $T$ NFEs. We note that by construction, for the forward process it holds that $\bm x_T \sim \mc N\pr{\bm 0, \bm I_N}$. This means that in the beginning of sampling, the latent variable $\bm x_T$ does not contain any information about $\bm x_0$. It is also reasonable to expect that a similar lack of information about $\bm x_0$ exists for a number of steps prior to $T$. Specifically, for those steps we expect that the quantity $\E\br{\bm x_0 |\bm y, \bm x_t}$ is heavily influenced by $\E\br{\bm x_0 |\bm y}$, while the contribution of $\E\br{\bm x_0 | \bm x_t}$ is not significant enough.  A theoretical justification for this argument is provided in~\cref{sec:proposition_proof}.

Based on the above reasoning, we select a timestep $\tau$ such that for the first $T - \tau$ reverse steps we use the following approximation: $\E\br{\bm x_0 | \bm y, \bm x_t} \approx \E\br{\bm x_0 | \bm y} = \bm \phi^{IR}_{\bm \theta_{IR}}\pr{\bm y}$. This is achieved by disabling the lower branch of our proposed conditional score matching network, namely the Denoising $\bm \phi^D_{\bm \theta_D}\pr{\tilde{\bm x}_t, \tilde{\sigma}_t}$ and Fusion $\bm \phi^F_{\bm \theta_F}\pr{\bm x_0^{IR}, \bm x_0^{D}}$ modules~(Figure~\ref{fig:denoiser_net}). Our strategy can be further supported by the recent study in~\cite{Choi2022perception}, where it has been demonstrated that the image sampling via DPMs could be divided into stages depending on the reverse process timesteps. In this spirit we activate the Denoising and Fusion modules at a timestep $\tau$ that is selected experimentally for the particular IR task of interest. Our results clearly indicate that the reconstruction result is going to be exactly the same whether we utilize the multi-step reverse diffusion process or the proposed one-step process that is described in Lemma~\ref{lem:single_step}. 

Indeed, since the quantity $\E\br{\bm x_0 | \bm y}$ is predicted by the IR network, which does not depend on the reverse diffusion parameters, $\bm \phi^{IR}_{\bm \theta_{IR}}\pr{\bm y}$ needs to be evaluated only once and its output can be re-used throughout the whole iterative sampling procedure. Expanding more on this idea, we show in Lemma~\ref{lem:single_step} that it is possible to omit entirely the first $T - \tau$ reverse diffusion steps and instead perform a single step directly from $T$ to $\tau$ with a procedure very similar to the one obtained for the diffusion process in~\cref{eq:cumulative_forward} from~\cref{eq:diffusion_kernel}. We provide the derivation in~\cref{sec:proposition_single_step}.

\begin{figure}[t]
  \begin{minipage}[t]{0.49\linewidth}
    \centering
    \label{fig:steps_skipped}
    \includegraphics[width=1\linewidth]{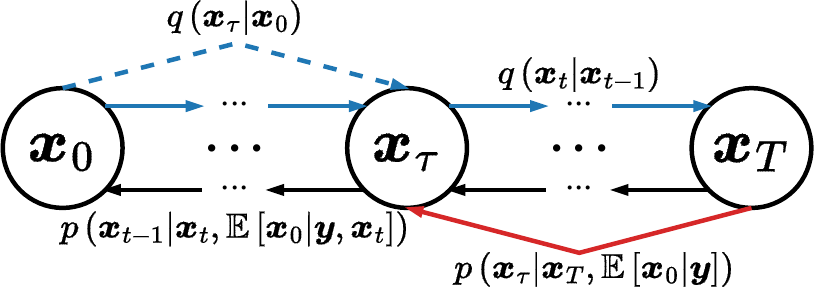}
    \vspace{-3mm}
    \caption{\small Forward and reverse diffusion process. Blue solid arrows: transitions at the forward pass with sampling distribution from~\cref{eq:diffusion_kernel}. Dashed arrow:  cumulative transition probability from~\cref{eq:cumulative_forward}. Black solid arrows: transitions at the backward pass with the sampling distribution from~\cref{eq:posterior}. Red solid arrow: closed-form cumulative transition probability from~\cref{eq:single_step} representing our accelerated sampling.}
  \end{minipage}%
  \hspace{0.01\linewidth} 
  \begin{minipage}[t]{0.49\linewidth}
    \scriptsize
    \centering
    \tabcolsep=0.05cm
    \captionof{table}{Performance evaluation on the task of Burst JDD-SR. We highlight the overall \glb{best} for each metric.}  
    \label{table:burst_results}
    \begin{tabular}{l|cc|cc|cc}
    \hline
    Methods    & PSNR\tts{$\uparrow$} & SSIM\tts{$\uparrow$}  & LPIPS\tts{$\downarrow$} & TOPIQ$_{\Delta}$\tts{$\downarrow$} & NFE \tts{$\downarrow$} & Params\tts{$\downarrow$} \\ \hline
    Target            & $\infty$    & 1           & 0            & 0           & N/A  & N/A   \\ \hline
    DBSR              & 31.98       & 0.891       & 0.198        & 0.10        & N/A  & 13.0M \\
    DeepRep           & 34.66       & 0.927       & 0.136        & 0.07        & N/A  & 12.1M \\
    EBSR              & 36.05       & 0.940       & 0.111        & 0.15        & N/A  & 26.0M \\
    BIPNet            & 34.86       & 0.934       & 0.112        & 0.03        & N/A  & 6.7M  \\
    BSRT-Small        & 35.91       & 0.940       & 0.109        & 0.12        & N/A  & 4.9M  \\
    BSRT-Large        & \glb{36.98} & \glb{0.947} & 0.095        & 0.16	       & N/A  & 20.7M \\ \hline
    Ours              & 35.53       & 0.933       & \glb{0.084}  & \glb{0.02}  & 6    & 21.6M \\ \hline
    \end{tabular}
  \end{minipage}
  \vspace{-3mm}
\end{figure}

\begin{lemma}\label{lem:single_step}
The transition probability defined in~\cref{eq:posterior} for a single reverse step, can be extended to $k$ reverse steps starting from $\bm x_t$ as:
\bal\label{eq:single_step}
p(\bm x_{t-k} | \bm x_t, \bm x_0) = \mc N \pr{\bm x_{t-k}; \bm \mu_{t,k}\pr{\bm x_t, \bm x_0}, {\sigma^2_{t, k}} \bm I_N},
\eal
where
\vspace{-2mm}
\bal\label{eq:single_step_meanvar}
&\bm \mu_{t,k}\pr{\bm x_t, \bm x_0} = \sum_{i=0}^{k-1} \Gamma_{t-i - 1}^{t-k+1} \gamma_{t-i} \bm x_0 + \Gamma_t^{t-k+1} \bm x_t \,\,\,\mbox{and}\,\,\,\sigma^2_{t,k} = \sum_{i=0}^{k-1} \pr{\Gamma_{t-i-1}^{t-k+1}}^2 \sigma_{t - i}^2.
\eal

In the above equations we make use of the following notation:
\bal\label{eq:single_step_gammas}
\gamma_t = \frac{\sqrt{\abar_{t-1}}\beta_t}{1 - \abar_t} \,\,\,\mbox{and}\,\,\, \Gamma_{i}^{j} \equiv 
\begin{cases}
\sqrt{\prod_{n=j}^i\alpha_n} \frac{1 - \abar_{j-1}}{1 - \abar_i} & \text{for } i \ge j\\ 
1 & \text{for } i < j
\end{cases} 
\eal

\end{lemma}

Since for the first $T - \tau$ steps $\bm x_0$ is approximated by the quantity $\E\br{\bm x_0 | \bm y}$, which is independent of the timestep $t$, we utilize~\cref{eq:single_step} to directly sample $\bm x_\tau$ as 
\bal\label{eq:non_standard_Gaussian}
\bm x_{\tau}& \sim p\pr{\bm x_\tau | \bm x_T, \E\br{\bm x_0 | \bm y}} =\mc N\pr{\bm x_\tau; \bm \mu_{T, T-\tau}\pr{\bm x_T, \E\br{\bm x_0 | \bm y}}, \sigma^2_{T, T-\tau}\bm I}.
\vspace{-1mm} 
\eal
This allows us to reduce the NFEs from $T\!+\!1$ to $\tau\!+\!1$, meaning that the required evaluations of Denoising + Fusion networks is reduced from $T$ to $\tau$. In both cases, we additionally count a single evaluation of the IR network. We depict our acceleration strategy with a red arrow in Figure 2. Finally, in practice we use $\tau = \tfrac{T}{200}$ for burst JDD-SR and dynamic scene deblurring, and $\tau = \tfrac{T}{4}$ for SISR, effectively reducing the NFEs by two orders of magnitude and a factor of four, respectively. \vspace{-1mm} 

The proposed acceleration procedure can be also interpreted as starting the sampling from step $\tau$ of the latent space using a non-standard Gaussian distribution as defined in~\cref{eq:non_standard_Gaussian}, instead of starting from step $T$ and using a standard Gaussian sample $\bm x_T \sim \mc N\pr{\bm 0, \bm I_N}$. We note, that a similar idea was explored in~\cite{Chung2022come,Meng2022sdedit}, where it was proposed to start the sampling from an observation that has been passed through a predefined number of forward diffusion steps. In our case the starting point for sampling is obtained via the approximated reverse process, which as a consequence of Lemma~\ref{lem:single_step} does not alter the final reconstruction result.
In another words, if we approximate $\boldsymbol{x}_0$ with $\mathbb{E}\left[\boldsymbol{x}_0|\boldsymbol{y}\right]$, then the reconstruction result will be the same both for the multi-step reverse diffusion process and the proposed one-step process. Moreover, it is easy to show that our strategy generalizes the one proposed in~\cite{Chung2022come,Meng2022sdedit}, as it leads to the same starting point if we make the following specific choices:
$\boldsymbol{x}_T\sim\mathcal{N}\left(\sqrt{\bar{\alpha}_T}\mathbb{E}\left[\boldsymbol{x}_0|\boldsymbol{y}\right],\left(1-\bar{\alpha}_T\right)\boldsymbol{I}\right)$ and $\sigma_t^2=\frac{1-\bar{\alpha}_{t-1}}{1-\bar{\alpha}_t}\beta_t$. In practice $\bar{\alpha}_T\approx 0$, so the first condition holds almost exactly. The second condition represents the particular choice of the noise variance used in the reverse process, with several existing parametrizations~\cite{ho2020denoising,nichol2021improved}. Our method is compatible with all of them and leads to different distributions for the starting point. In all our experiments we use the parametrization $\sigma_t^2=\beta_t$ from~\cite{ho2020denoising}. More detailed conceptual and technical differences along with experimental results are provided in~\cref{sec:appendix_acceleration_difference}.
 
Furthermore, our acceleration strategy is complimentary to other sampling acceleration techniques~\cite{song2021denoising,karras2022elucidating}. To demonstrate this, we utilize the DDIM~\cite{song2021denoising} sampling to further reduce the NFEs by a factor of five for the SISR task. As a result, the reverse diffusion process used for this problem requires $\tfrac{T}{20} + 1$ NFEs in total.
\vspace{-3mm} 
\section{Experimental Results}
\label{sec:exps}
\vspace{-2mm} 
We evaluate our method on four public datasets across a range of tasks, namely burst JDD-SR, dynamic scene deblurring, and SISR. Below we describe our specific architecture and design choices related to all utilized modules. 
\vspace{-2mm} 

\paragraph{Training.} Our training procedure consists of two stages. We first employ a diverse, yet small DF2K (combination of DIV2K~\cite{agustsson2017ntire} and Flickr2K~\cite{lim2017enhanced}) dataset to train a Denoising Module for Gaussian denoising in the sRGB domain with input noise levels ranging in $[0, 244.3]$. These noise levels corresponds to timesteps in the range of $[0, 250]$ for the diffusion process with $T = 1000$. We use the original 
training procedure of MIRNet~\cite{zamir2020learning} to learn the parameters of our MIRNet-S architecture. At the second stage we train our Fusion Modules with $n_f = 64$ for each IR task and the corresponding pre-trained off-the-shelf IR network. It is worth noting, that at this stage the parameters of Denoising and IR modules are kept frozen and only the Fusion Module is trained. Specifically, we train it for $300k$ iterations with a learning rate of $10^{-4} \times 0.99$ it/1000, batch size of 128, and crop size of $256 \times 256$. 
To train our Fusion networks we use datasets that are common among our main competitors, specifically the ZurichRAW2RGB~\cite{ignatov2020replacing} dataset for burst JDD-SR, GoPro~\cite{nah2017deep} for dynamic scene deblurring and DIV2K~\cite{agustsson2017ntire} for $4\times$ SISR. For burst JDD-SR the Fusion network is trained in the sRGB domain. All the networks are trained using the Ascend 910 AI accelerators~\cite{liao2021ascend}. To make our results reproducible, we provide a full description of the training procedure in~\cref{sec:supp_training}. 
\vspace{-2mm} 

\paragraph{Inference.}
For each IR task we use the procedure described in~\cref{sec:accelerated_sampling} to obtain the reconstructed images with $T = 1000$. To demonstrate the effectiveness of our approach, for each problem of interest except for burst JDD-SR we need half of the NFEs compared to the diffusion-based competitor that uses the least number of sampling steps. Specifically, for dynamic scene deblurring we use $\tau = 5$, resulting in $200\times$ acceleration achieved solely by our proposed sampling strategy. This amounts to $6$ NFEs when counting the additional IR network evaluation performed to skip the first $T - \tau = 995$ steps using~\cref{eq:single_step}. For SISR we select $\tau = 250$, which results in $4\times$ acceleration using our sampling procedure. In order to demonstrate how it can be complemented by other proposed acceleration strategies, for the final $\tau=250$ steps we achieve $5\times$ step reduction by employing DDIM sampling~\cite{song2021denoising}. The combination of both acceleration strategies results in $20\times$ step reduction and $51$ NFEs overall. Applying the DDIM acceleration technique on top of our proposed one-step strategy leads to an insignificant quantitative/qualitative difference (see~\cref{sec:appendix_ddim_combo}) compared to our original scheme. Since for the burst JDD-SR problem no diffusion-based methods have yet been proposed, we use the same setting as for the dynamic scene deblurring problem, as it requires the smallest NFEs. In all our experiments we use the linear scheduling of the diffusion process variances $\beta_t\in \br{2\times10^{-2}, 10^{-4}}$ defined in~\cref{eq:diffusion_kernel}. 

\paragraph{Evaluation.} \vspace{-2mm} 
For the burst JDD-SR evaluation we use the SyntheticBurst test set~\cite{dbsr21}, consisting of 300 synthetically pre-generated raw burst sequences. Each sequence contains 14 noisy raw LR images with handshake motion, whose corresponding targets have a resolution of $320 \times 320$. Since our networks are trained on sRGB images, the outputs of all methods are converted to the sRGB space prior to comparison. For dynamic scene deblurring we evaluate on the GoPro test~\cite{nah2017deep} and HIDE~\cite{shen2019human} benchmarks, which contain 1111 and 2025 images of 720p resolution, respectively. For $4\times$ SISR we use the DIV2K validation dataset~\cite{agustsson2017ntire} consisting of 100 images of 2K resolution. \vspace{-0.5mm}

For the quantitative evaluation of the reconstruction quality we rely on the widely used fidelity metrics PSNR and SSIM~\cite{wang2004image}, and the reference-based perceptual metric LPIPS~\cite{zhang2018perceptual}. Moreover, we also utilize the non-reference image quality assessment (NR-IQA) metric TOPIQ~\cite{chen2023topiq} 
and report the absolute distance between the output and the target scores, which we indicate as TOPIQ$_\Delta$.

\begin{table}[!t]
\vspace{-2mm}
\caption{Performance evaluation on the GoPro and HIDE test sets for dynamic scene deblurring. $^\dagger$ indicates that public implementation is unavailable and the scores are copied from the authors' paper. We highlight the overall \glb{best} for each metric, and the \lcl{best} among perceptual-oriented methods.}
\scriptsize
\centering
\tabcolsep=0.05cm
\label{table:deblur_mix}
\begin{tabular}{l|cccc|cccc|cc}
\hline
\multirow{2}{*}{Methods} & \multicolumn{4}{c|}{GoPro} & \multicolumn{4}{c|}{HIDE} & \multirow{2}{*}{NFE \tts{$\downarrow$}} & \multirow{2}{*}{Params \tts{$\downarrow$}}\\

& PSNR\tts{$\uparrow$} & SSIM\tts{$\uparrow$}  & LPIPS\tts{$\downarrow$} & TOPIQ$_{\Delta}$\tts{$\downarrow$} & PSNR\tts{$\uparrow$} & SSIM\tts{$\uparrow$}  & LPIPS\tts{$\downarrow$} & TOPIQ$_{\Delta}$\tts{$\downarrow$} \\ \hline
Target              & $\infty$    & 1           & 0            & 0            & $\infty$          & 1           & 0           & 0           & N/A  & N/A   \\ \hline
HINet               & 32.77       & 0.960       & 0.088        & 0.033        & 30.33             & 0.932       & 0.120       & 0.044       & N/A  & 88.6M \\
MPRNet              & 32.66       & 0.959       & 0.089        & 0.027        & 30.96             & 0.939       & 0.114       & 0.059       & N/A  & 20.1M \\
MIMO-UNet+          & 32.44       & 0.957       & 0.091        & 0.034        & 29.99             & 0.930       & 0.124       & 0.028       & N/A  & 16.1M \\
NAFNet              & 33.71       & 0.967       & 0.078        & 0.017        & 31.32             & 0.943       & 0.103       & 0.024       & N/A  & 67.9M \\
Restormer           & 32.90       & 0.961       & 0.084        & 0.018        & 31.20             & 0.942       & 0.109       & 0.048       & N/A  & 26.1M \\
FFTFormer           & \glb{34.21} & \glb{0.969} & 0.071        & 0.012        & \glb{31.62}       & \glb{0.946} & 0.096       & 0.006       & N/A  & 16.6M \\ \hline
\multicolumn{11}{c}{Perceptual-oriented Methods}                                                                                                            \\ \hline      
DeblurGANv2         & 29.08       & 0.918       & 0.117        & 0.025        & 27.51             & 0.885       & 0.159       & 0.065       & N/A  & 60.9M \\
DvSR$^\dagger$      & 31.66       & 0.948       & 0.059        & -            & 29.77             & 0.922       & 0.089       & -           & 500  & 26.1M \\
icDPM$^\dagger$     & 31.19       & 0.943       & 0.057        & -            & 29.14             & 0.910       & 0.088       & -           & 500  & 52.0M \\
InDi$^\dagger$      & 31.49       & 0.946       & 0.058        & -            & \multicolumn{4}{c|}{-}                                      & 10   & 27.7M \\
Ours                & \lcl{33.72} & \lcl{0.963} & \glb{0.053}  & \glb{0.011}  & \lcl{31.32}       & \lcl{0.937} & \glb{0.087} & \glb{0.002} & 6    & 33.2M \\ \hline
\end{tabular}
\vspace{-1.5mm} 
\end{table}

{\subsection{Results}}

\begin{figure*}[t]
\centering
\setlength\extrarowheight{-10pt}
\begin{tabular}{@{\hskip 0.05cm} c @{\hskip 0.05cm} c @{\hskip 0.05cm} c @{\hskip 0.05cm} c @{\hskip 0.05cm} c @{\hskip 0.05cm} c @{\hskip 0.05cm}}
  \multirow{3}{*}[1.64cm]{\includegraphics[width=.234\linewidth]{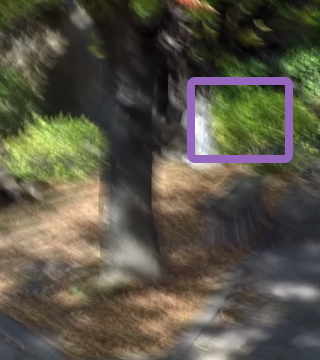}} &
 \includegraphics[width=.151\linewidth]{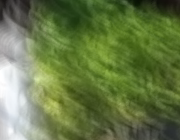} & 
 \includegraphics[width=.151\linewidth]{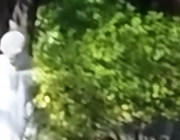} & 
 \includegraphics[width=.151\linewidth]{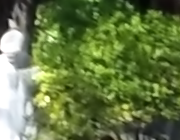} &
 \includegraphics[width=.151\linewidth]{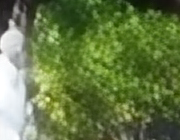} &
 \includegraphics[width=.151\linewidth]{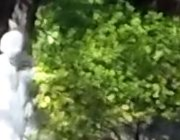}\\
 & \scriptsize{Input: 0.424} & \scriptsize{HINet: 0.127} & \scriptsize{Restormer: 0.136} & \scriptsize{DeblurGANv2: 0.171} & \scriptsize{InDI: 0.080} \\
 
 & \includegraphics[width=.151\linewidth]{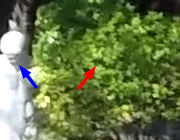} & 
   \includegraphics[width=.151\linewidth]{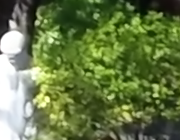} & 
   \includegraphics[width=.151\linewidth]{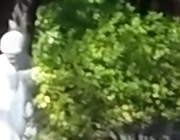} &
   \includegraphics[width=.151\linewidth]{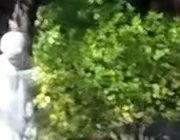} &
   \includegraphics[width=.151\linewidth]{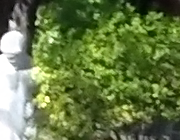} \\
  & \scriptsize{Target} & \scriptsize{NAFNet: 0.083} & \scriptsize{FFTFormer: \textbf{0.059}} & \scriptsize{DvSR: 0.072} & \scriptsize{Ours: \textbf{0.059}}
\end{tabular}
\vspace{-2mm}
   \caption{Visual comparisons on the GoPro test set for the task of dynamic scene deblurring (best viewed by zooming in). Every output image is accompanied by its LPIPS value.}
   \vspace{-2mm}
\label{fig:gopro_deblur}
\end{figure*}

\paragraph{Burst JDD-SR.} 
In Table~\ref{table:burst_results} we compare our proposed pipeline with existing methods, namely DBSR~\cite{dbsr21}, DeepRep~\cite{deeprep21}, and the current SOTA methods, namely BIPNet~\cite{dudhane2022burst}, BSRT-Small~\cite{luo2022bsrt}, BSRT-Large~\cite{luo2022bsrt}, and  EBSR~\cite{luo2021ebsr}. Our method demonstrates SOTA performance across the perceptual metrics while maintaining competitive PSNR and SSIM scores compared to existing methods. Thus, our method reconstructs images that are closer to the target based on the human perception while maintaining a high level of fidelity. We refer to figures in the~\cref{sec:supp_addition_imgs} for a qualitative visual assessment. Furthermore, we notice an improvement in terms of visual quality and perceptual metrics compared to BSRT-Small, which we use as the IR module of choice in our framework. This indicates that our approach preserves the fidelity of the IR model outputs, while enhancing their perceptual quality by running few reverse diffusion steps. It is worth noting that for this case, where a burst of raw images serves as input, we use the exact same denoising network that was trained on sRGB images and which we later deploy to all considered single-input IR tasks. This highlights the generalization ability of our approach not only to different IR problems but also to different input formats. Note that AFCNet~\cite{mehta2022adaptive}, LKR~\cite{lecouat2021lucas}, and Burstormer~\cite{dudhane2022burst} are not included in our comparisons due to the absence of a publicly available implementation (or trained network parameters). Finally, the comparison with SOTA EBSR and BSRT shows that our DP-IR reconstructions compares favorably in terms of visual quality, while not lacking significantly in terms of fidelity. 
\vspace{-2mm}
\paragraph{Dynamic Scene Deblurring.}

Table~\ref{table:deblur_mix} show quantitative results on the GoPro~\cite{nah2017deep} and HIDE~\cite{shen2019human} datasets, respectively. We compare our  approach with the SOTA reconstruction-based methods: NAFNet~\cite{ren2023multiscale}, FFTFormer~\cite{kong2023efficient} and diffusion-based methods: DvSR~\cite{whang2022deblurring}, InDI~\cite{delbracio2023inversion}, and icDPM~\cite{ren2023multiscale}. Our framework outperforms all competing methods across the perceptual metrics and demonstrates the best perception-distortion (P-D) trade-off among all perceptual-based methods. Moreover, our DP-IR uses twice less number of reverse steps (NFE=5) compared to the state-of-the-art InDI and still achieves better perceptual quality (e.g. LPIPS) and is more consistent with the ground-truth (+2.22dB in PSNR). Despite the fact that our fusion module is trained only on the GoPro dataset, the gains in perceptual quality do transfer over the HIDE test images, providing SOTA scores for the perceptual metrics.
Also, among perceptual-oriented methods, DP-IR outperforms the closest competitor DvSR by 1.55dB in terms of PSNR. Visual comparisons of our method and the SOTA deblurring models: NAFNet, FFTFormer, DvSR, and InDI is depicted in Figure~\ref{fig:gopro_deblur}. From these results we observe that our model shows a noticeable improvement in perceptual quality. Moreover, we have performed a computational cost analysis for the diffusion-based models and significantly outperformed existing methods from $\sim$2 to 100 times  (see~\cref{sec:appendix_complexity}). 
\begin{figure*}[t]
\vspace{-4mm}
\centering
\setlength\extrarowheight{-10pt}
\begin{tabular}{@{\hskip 0.05cm} c @{\hskip 0.05cm} c @{\hskip 0.05cm} c @{\hskip 0.05cm} c @{\hskip 0.05cm} c @{\hskip 0.05cm} c @{\hskip 0.05cm}}
  \multirow{3}{*}[1.74cm]{\includegraphics[width=.179\linewidth]{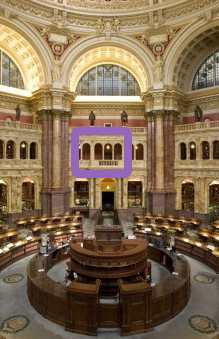}} &
 \includegraphics[width=.16\linewidth]{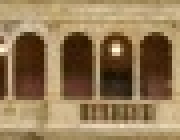} & 
 \includegraphics[width=.16\linewidth]{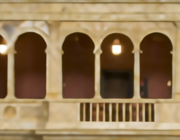} & 
 \includegraphics[width=.16\linewidth]{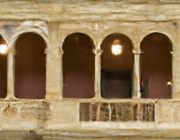} &
 \includegraphics[width=.16\linewidth]{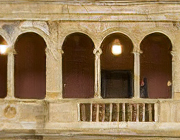} &
 \includegraphics[width=.16\linewidth]{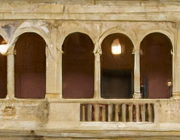}\\
 & \scriptsize{Input: 0.341} & \scriptsize{SwinIR: 0.123} & \scriptsize{HCFlow: 0.109} & \scriptsize{ESRGAN: 0.092} & \scriptsize{InDI: 0.078} \\
 
 & \includegraphics[width=.16\linewidth]{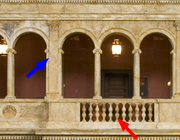} & 
 \includegraphics[width=.16\linewidth]{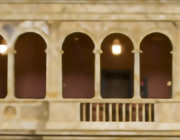} & 
 \includegraphics[width=.16\linewidth]{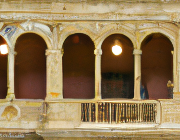} &
 \includegraphics[width=.16\linewidth]{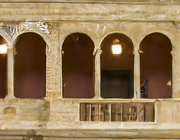} &
 \includegraphics[width=.16\linewidth]{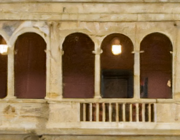} \\
  & \scriptsize{Target} & \scriptsize{HAT: 0.128} & \scriptsize{LDM: 0.140} & \scriptsize{SRDiff: 0.122} & \scriptsize{Ours: \textbf{0.077}}
\end{tabular}
\vspace{-2mm}
   \caption{Visual comparisons on the DIV2K validation set for the task of $4\times$ bicubic super-resolution (best viewed by zooming in). Every output image is accompanied by its LPIPS value.}
\label{fig:div2k_images}
\vspace{-3mm}
\end{figure*}

\begin{table}[!t]
    \scriptsize
    \centering
    \tabcolsep=0.03cm
    \begin{minipage}[t]{.49\linewidth}
      \centering
      \caption{\small Performance evaluation on the DIV2K validation set for $4\times$ SISR. We highlight the overall \glb{best} for each metric, and the \lcl{best} among perceptual-oriented methods.}
      \label{table:div2k_results}
        \begin{tabular}{l|cc|cc|cc}
        \hline 
        Methods    & PSNR\tts{$\uparrow$} & SSIM\tts{$\uparrow$}  & LPIPS\tts{$\downarrow$} & TOPIQ$_{\Delta}$\tts{$\downarrow$} & NFE \tts{$\downarrow$} & Params\tts{$\downarrow$} \\ \hline
        Target       & $\infty$       & 1            & 0            & 0            & N/A   & N/A \\ \hline
        SRResNet     & 29.07          & 0.824        & 0.266        & 0.046        & N/A   & 1.5M \\
        RRDB         & 29.48          & 0.834        & 0.254        & 0.038        & N/A   & 16.7M \\
        SwinIR       & 29.63          & 0.837        & 0.248        & 0.030        & N/A   & 11.9M \\
        LIIF         & 29.30          & 0.830        & 0.258        & 0.046        & N/A   & 22.3M \\
        HAT          & \glb{29.75}    & \glb{0.840}  & 0.245        & 0.035        & N/A   & 20.6M \\ \hline
        \multicolumn{7}{c}{Perceptual-oriented Methods} \\                                       \hline
        ESRGAN       & 26.64           & 0.758        & \glb{0.115} & 0.014        & N/A   & 16.7M \\
        HCFlow       & 27.02          & 0.766        & 0.124        & 0.021        & N/A   & 23.2M \\ 
        SwinIR-GAN   & 24.88          & 0.734        & 0.222        & 0.115        & N/A   & 11.9M \\ 
        LDM          & 23.30          & 0.697        & 0.218        & 0.019        & 100   & 169.0M \\
        SRDiff       & 27.14          & 0.773        & 0.129        & 0.008        & 100   & 23.6M \\
        InDI         & 26.45          & 0.741        & 0.136        & 0.009        & 100   & 62.3M \\   
        IDM          & 27.35          & 0.782        & 0.147        & 0.008        & 2000  & 116.6M \\
        Ours         & \lcl{28.12}    & \lcl{0.793}  & 0.140        & \glb{0.002}  & 51    & 28.5M \\ \hline
        \end{tabular}
    \end{minipage}%
    \hfill
    \begin{minipage}[t]{.49\linewidth}
        \centering
        \caption{\small Ablation on various denoiser and IR networks on DIV2K validation for $4\times$ SISR.}
        \label{table:ablation_plug_play}
            \begin{tabular}{ll|cc|cc}
                \hline 
                Denoiser & IR Network    & PSNR\tts{$\uparrow$} & SSIM\tts{$\uparrow$}  & LPIPS\tts{$\downarrow$} & TOPIQ$_{\Delta}$\tts{$\downarrow$} \\ \hline
                \multicolumn{2}{c|}{Target}          & inf            & 1            & 0            & 0            \\ \hline
                UDP                              & RRDB   & 27.93          & 0.777        & 0.149        & 0.006        \\ 
                MIRNet-S                         & RRDB   & 28.12          & 0.795        & 0.150        & 0.014        \\ 
                MIRNet-S                         & SwinIR & 28.12          & 0.793        & 0.140        & 0.002        \\ \hline
            \end{tabular}
        \vspace{-1mm}
        \caption{\small Ablation on various fusion networks on DIV2K validation for $4\times$ SR task}
        \label{table:ablation_fuser}
            \begin{tabular}{l|ccc|c}
                \hline 
                Fusion Module                    & PSNR$\uparrow$ & SSIM$\uparrow$  & LPIPS$\downarrow$ & Params \\ \hline
                TDWA                             & 27.26          & 0.751           & 0.266             & 8385   \\ 
                L-DWT                            & 29.27	      & 0.818 	        & 0.233             & 30.0M  \\
                U-Net                            & 28.19          & 0.785           & 0.141             & 31.0M  \\ 
                Proposed Fusion                  & 27.93          & 0.777           & 0.149             & 0.7M   \\ \hline
            \end{tabular}
        \vspace{-2mm}
        \centering
        \caption{\small Perception-Distortion trade-off on DIV2K validation for $4\times$ SISR.}
        \label{table:ablation_tau}
            \scriptsize
            \begin{tabular}{l|ccccccc}
                \hline
                Timestep, $\tau$   & 50          & 100   & 150   & 200   & 250    & 300    & 350         \\ \hline
                PSNR $\uparrow$    & \glb{28.77} & 28.46 & 28.24 & 28.06 & 27.93  & 27.81  & 27.71       \\
                LPIPS$\downarrow$  & 0.170       & 0.161 & 0.155 & 0.152 & 0.149  & 0.147  & \glb{0.146} \\ \hline 
            \end{tabular}
    \end{minipage} 
    \vspace{-5mm}
\end{table}

\vspace{-2mm}
\paragraph{Super-Resolution.} We compare our method with reconstruction-based~\cite{ledig2017photo, wang2018esrgan, liang2021swinir, chen2021learning, chen2023activating}, GAN-based~\cite{wang2018esrgan, zhang2021ranksrgan, liang2021swinir}, NF-based~\cite{liang2021hierarchical} models, and DPMs~\cite{rombach2022high, li2021srdiff, delbracio2023inversion, gao2023implicit}.  Table~\ref{table:div2k_results} summarizes the quantitative results on the DIV2K validation set. Our solution produces the best fidelity scores among all six perceptual-based methods and the best TOPIQ perceptual metric among all competing methods. 
The visual comparison in Figure~\ref{fig:div2k_images} reveals that our framework produces super-resolved images that exhibit more refined structures and fine-grained details.
Additional examples for all reported IR tasks are provided in~\cref{sec:supp_addition_imgs}.

\vspace{-2mm}
\section{Ablation Studies} \label{sec:ablation}
\paragraph{Modular Approach.} \label{subsec:ablation_modular} 
One of the main benefits of our framework is its ability to capitalize on the performance of existing restoration networks at a relatively low additional computational cost. To showcase this, we conduct an experiment on the task of $4\times$ SISR using two different denoising architectures, namely UDP~\cite{brooks2019unprocessing} and MIRNet-S~\cite{zamir2020learning}, and two IR architectures, namely RRDB~\cite{agustsson2017ntire} and SwinIR~\cite{liang2021swinir}. UDP is trained with the same settings as MIRNet-S and the fusion module is retrained for each of the three cases. From Table~\ref{table:ablation_plug_play}, we observe that the same IR network combined with a better denoising module, leads to better fidelity (see PSNR, SSIM). A same trend, but for perceptual metrics is observed if one upgrades the IR module from RRDB to the SwinIR and keeps the same denoising module. This clearly indicates that one can achieve better results by employing either a more powerful denoiser or IR module, without the need to fully retrain the entire score estimator as is the common practice followed in most of the existing methods (\eg LDM, SRDiff, DvSR, \etc).
\vspace{-2mm}
\paragraph{Fusion Strategies.} In this study we use the UDP denoiser and the RRDB network from Table~\ref{table:ablation_plug_play} and study several different fusion approaches, namely the Time-dependent Weighted Averaging (TDWA), Learnable Discrete Wavelet Transform (L-DWT), our proposed Fusion network, and U-Net with time embeddings~\cite{pmlr-v139-nichol21a}. TDWA consists of sinusoidal positional encoding followed by a three-layer MLP, which predicts the weights for the timestep $t$. 

Those weights are then passed to the 3CA layer (\cref{sec:network_architecture}) together with the outputs from the denoiser and IR modules to perform the fusion in the image space.
In contrast, L-DWT directly learns weights for each scale and channel of a 3-level Haar DWT~\cite{haar1909theorie}. L-DWT has only 30 trainable parameters, which needs significantly less training time and data. Table~\ref{table:ablation_fuser} indicates that, as expected, a more powerful fusion module leads to better perceptual and reconstruction quality. Overall, we see that the proposed Fusion network shows a better balance between the reconstruction, visual quality and the computational cost.    

\vspace{-2mm}
\paragraph{Perception-Distortion Trade-off.} By varying the timestep $\tau$, when the denoiser and fusion modules are activated, one can favor the perceptual quality over the reconstruction fidelity (see Table~\ref{table:ablation_tau}). Here, we use the same UDP denoiser and RRDB as the IR module and experiment on the DIV2K~\cite{agustsson2017ntire} validation for the task of $4\times$ SISR. Table~\ref{table:ablation_tau} shows that the denoiser can operate on $\tau > 250$, which corresponds to a wider noise range than the one the denoiser is initially trained for. Furthermore, we observe the same perception-distortion trade-off for dynamic scene deblurring task (see Appendix Table~\ref{table:ablation_tau_gopro})

\vspace{-2mm}
\paragraph{Limitations}\label{sec:limitations}

Our empirical findings highlight that the optimal selection of $\tau$ is intrinsically linked to the nature of the reconstruction problem, particularly the output quality of the IR Network. While we have experimentally identified optimal parameters for each test dataset in this study, we posit that a more refined approach would involve tailoring the acceleration parameters on an individual sample basis. However, the absence of a dependable methodology for assessing the quality of the IR and Denoising Networks' outputs at specific diffusion process timesteps – especially in the absence of ground truth data – constitutes a considerable challenge. This underscores a compelling avenue for future inquiry into adaptive optimization of acceleration parameters.

Furthermore, a notable constraint of our approach is its reliance on the efficacy of the employed Denoising and IR modules. As such, for novel image restoration tasks where a pre-trained IR network is unavailable, our framework might be inapplicable. Additionally, for imaging modalities (e.g. medical imaging) lacking a trained score-matching network (denoising module), it is imperative to either fine-tune an existing module or undertake comprehensive re-training with appropriate image datasets.

\vspace{-4mm}
\section{Conclusion}
\vspace{-3mm}
\label{sec:conclusion}
We present a modular conditional diffusion probabilistic framework for IR problems along with a sampling acceleration strategy that achieves a significant speed-up during the inference stage. Our framework achieves SOTA results both quantitatively and visually on the tasks of burst JDD-SR, dynamic scene deblurring, and $4\times$ SISR without the need for re-training on a large pool of data and significant computational cost. This is mainly accomplished by utilizing pretrained models and only training a relatively small fusion module. While in this work we have not exhaustively considered all blind IR problems, we hope that our results can serve as a positive indication that the perceptual quality of the reconstructed outputs can improve significantly at the cost of only several additional NFEs, making possible a wider adoption of DPMs for IR applications even when there are tight requirements on computational complexity. Our ablation studies indicate that a variety of pretrained networks can be used with our method and further improvements on the results can be achieved by utilizing better denoising, IR, and fusion modules.

\bibliography{main}
\bibliographystyle{cvpr_bib}

\appendix
\newpage
\section{Conditional DPM Training Requirements}\label{sec:appendix_dpm_requirements}
To the best of our knowledge, all diffusion-based approaches that have been proposed in the literature to deal with image restoration tasks, require the training of far larger conditional backbone networks ($\sim$10-100M params). This turns out to be significantly more challenging both in terms of necessary training data and computational resources. To showcase this, we provide an indicative example below. If we adopt the existing diffusion-based SISR baselines and train them for a completely different restoration problem, by following the original authors' training strategies it turns out that the computational and data requirements are significantly higher than those of our method.

\begin{table}[!h]
\centering
\caption{Comparison of the proposed approach against existing DPM methods for SISR task in terms of training dataset requirements and training parameters.}
\label{table:dpm_requirements}
\begin{tabular}{l|c|c} 
\hline
Method & Params required & Data required \\ \hline 
Ours   & $1$x & $1$x \\
SRDiff & $\sim 34$x & $\sim 4$x \\
LDM    & $\sim 240$x & $\sim 1000$x \\
InDI   & $\sim 89$x & $\sim 1$x \\
IDM    & $\sim 167$x & $\sim 1$x \\
\hline
\end{tabular}
\end{table}

Based on these data, we can safely state that our strategy provides a reasonable trade-off between the required training complexity and the competitive performance of our method to a variety of blind inverse problems.
 
\section{Proof of Lemma~\ref{lem:score_function}}\label{sec:score_function_proof}
To derive the conditional score function, we first express the conditional probability $p\pr{\bm x_t | \bm y}$ as: 
\bal
p\pr{\bm x_t | \bm y} = \int p\pr{\bm x_t, \bm x_0 | \bm y} \,d\bm x_0 = \int p\pr{\bm x_t |\bm y, \bm x_0} p\pr{\bm x_0 |\bm y} \,d\bm x_0 =\int q\pr{\bm x_t | \bm x_0} p\pr{\bm x_0 |\bm y} \,d\bm x_0,
\label{eq:supp_conditional_marginal}
\eal
where we used the fact that $\bm x_t$ is conditionally independent of $\bm y$ and according to~\cref{eq:cumulative_forward} it holds $p\pr{\bm x_t |\bm y, \bm x_0}=q\pr{\bm x_t | \bm x_0}$. Differentiating both sides of~\cref{eq:supp_conditional_marginal} with respect to (w.r.t.) $\bm x_t$ we get:
\bal
\grad{p\pr{\bm x_t | \bm y}}{\bm x_t} = \int p\pr{\bm x_0 |\bm y} \grad{q\pr{\bm x_t | \bm x_0}}{\bm x_t}  \,d\bm x_0.
\label{eq:grad_cprob_start}
\eal
Based on the definition of $q\pr{\bm x_t | \bm x_0}$ in~\cref{eq:cumulative_forward}, it also holds that: $\grad{q\pr{\bm x_t | \bm x_0}}{\bm x_t} = -q\pr{\bm x_t | \bm x_0} \frac{\bm x_t - \sqrt{\abar_t} \bm x_0}{1 - \abar_t}$. Substituting this result back to~\cref{eq:grad_cprob_start}, we get:
\bal
\grad{p\pr{\bm x_t | \bm y}}{\bm x_t} &= \frac{\sqrt{\abar_t}}{1 - \abar_t} \int \bm x_0 q\pr{\bm x_t | \bm x_0} p\pr{\bm x_0 |\bm y} \, d\bm x_0
-\frac{\bm x_t}{1 - \abar_t}\overbrace{\int q\pr{\bm x_t | \bm x_0} p\pr{\bm x_0 |\bm y} \, d\bm x_0}^{p\pr{\bm x_t | \bm y}}.
\label{eq:eq:grad_cprob}
\eal

\noindent Next, if we divide both sides in~\cref{eq:grad_cprob_start} with $p\pr{\bm x_t | \bm y}$ and use that $\grad{\log{p\pr{\bm x_t | \bm y}}}{\bm x_t} = \frac{\grad{p\pr{\bm x_t | \bm y}}{\bm x_t}}{p\pr{\bm x_t | \bm y}}$, we get:
\bal
\grad{\log{p\pr{\bm x_t | \bm y}}}{\bm x_t} = \frac{1}{\pr{1 - \abar_t}}\pr{\sqrt{\abar_t}\int \bm x_0 \frac{q\pr{\bm x_t | \bm x_0} p\pr{\bm x_0 |\bm y}}{p\pr{\bm x_t | \bm y}} \, d\bm x_0 - \bm x_t}.
\label{eq:supp_conditional_score_fn}
\eal
We can further express the integral in~\cref{eq:supp_conditional_score_fn} as follows:
\bal
\nonumber&\int \bm x_0 \frac{q\pr{\bm x_t | \bm x_0} p\pr{\bm x_0 |\bm y}}{p\pr{\bm x_t | \bm y}} \, d\bm x_0 = \int \bm x_0 \frac{q\pr{\bm x_t | \bm x_0} p\pr{\bm x_0 |\bm y}p\pr{\bm y}}{p\pr{\bm x_t | \bm y}p\pr{\bm y}} \, d\bm x_0 = \int \bm x_0 \frac{p\pr{\bm x_t | \bm x_0, \bm y} p\pr{\bm x_0 , \bm y}}{p\pr{\bm x_t, \bm y}} \, d\bm x_0 \\\nonumber 
= &\int \bm x_0 \frac{p\pr{\bm x_t, \bm x_0, \bm y}}{p\pr{\bm x_t, \bm y}} \, d\bm x_0 = \int \bm x_0 p\pr{\bm x_0 | \bm y, \bm x_t} \, d\bm x_0 = \E \br{\bm x_0 | \bm y, \bm x_t}.
\eal
Substituting this result in~\cref{eq:supp_conditional_score_fn} finally leads us to the result of the lemma: $\grad{p\pr{\bm x_t | \bm y}}{\bm x_t} = \frac{\sqrt{\abar_t}\E \br{\bm x_0 | \bm y, \bm x_t} - \bm x_t}{1 - \abar_t}$. 

\section{Theoretical Justification of the Conditional Expectation Approximation}\label{sec:proposition_proof}
By construction, for the forward process from~\cref{eq:diffusion_kernel} it holds that $\bm x_T \sim \mc N\pr{\bm 0, \bm I_N}$ when $T\rightarrow \infty$, which means that in the beginning of reverse sampling, the latent variable $\bm x_T$ does not contain any information about $\bm x_0$. Below we provide a theoretical result that can serve as an indication that during the sampling process and up to some timestep $\tau$, we can use an approximation of $\E\br{\bm x_0 |\bm y, \bm x_t} \approx \E\br{\bm x_0 |\bm y}$, given that the contribution of $\E\br{\bm x_0 | \bm x_t}$ is not significant enough. This is achieved by disabling the lower branch of our proposed conditional score matching network, which includes the Denoising $\bm \phi^D_{\bm \theta_D}\pr{\tilde{\bm x}_t, \tilde{\sigma}_t}$ and Fusion $\bm \phi^F_{\bm \theta_F}\pr{\bm x_0^{IR}, \bm x_0^{D}}$ modules~(Figure~\ref{fig:denoiser_net}). A formal theoretical analysis is available under the following simplifications:
\begin{itemize}
    \item The observation model is linear: $\bm y = \bm A \bm x_0 + \bm n$.
    \item Denoising Network is an identity transformation:  $\boldsymbol{\phi}^D_{\boldsymbol{\theta}_D}\left(\tilde{\boldsymbol{x}}_t, \tilde{\sigma}_t\right) \equiv \tilde{\boldsymbol{x}}_t$.
    \item Image Restoration Network is a back-projection: $\boldsymbol{\phi}^{IR}_{\boldsymbol{\theta}_{IR}}\left(\boldsymbol{y}\right) \equiv \boldsymbol{A}^T\boldsymbol{y}$
    \item Fusion Network is a convex combination in a spatial domain: $\boldsymbol{\phi}^F_{\boldsymbol{\theta}_F}\left(\boldsymbol{x}_0^{IR}, \boldsymbol{x}_0^{D}, t\right) \equiv w\boldsymbol{x}_0^{IR} + (1-w) \boldsymbol{x}_0^{D}$.
    \item Diffusion process approaches the continuous-time regime: $T \rightarrow \infty$.
\end{itemize}
Our theoretical result is provided in the form of the following proposition.

\begin{proposition}\label{prop:latent_observation_comparison}
Let $\boldsymbol{y}\in\mathbb{R}^M$ be the measurements obtained according to the following observation model: $\boldsymbol{y} = \boldsymbol{A}\boldsymbol{x_0}+\boldsymbol{n}$, where $\boldsymbol{A}\in\mathbb{R}^{M \times N},\lVert\boldsymbol{A}\rVert_2\le1$ and $\boldsymbol{n}\sim\mathcal{N}\left(\boldsymbol{0},\sigma_{\boldsymbol{y}}\boldsymbol{I}_M\right),\sigma_{\boldsymbol{y}}\le1$. Then, for any given $\boldsymbol{A}$,$\sigma_{\boldsymbol{y}}$,$\boldsymbol{x}_0\in\mathbb{R}^N$, and any fixed $w\in(0,1)$, there exists a timestep $\tau$, such that for all $t > \tau$ the back-projected signal $\boldsymbol{A}^T\boldsymbol{y}$ approximates $\boldsymbol{x}_0$ better than the convex combination $w\boldsymbol{A}^T\boldsymbol{y}+(1-w)\tilde{\boldsymbol{x}}_t$ in the following sense: 
\bal\label{eq:proposition}
\mathbb{E}_{\boldsymbol{y}|\boldsymbol{x}_0}\lVert\boldsymbol{x}_0-\boldsymbol{A}^T\boldsymbol{y}\rVert_2^2\le\mathbb{E}_{\boldsymbol{y},\boldsymbol{x}_t|\boldsymbol{x}_0}\lVert\boldsymbol{x}_0-\left(w\boldsymbol{A}^T\boldsymbol{y}+(1-w)\tilde{\boldsymbol{x}}_t\right)\rVert_2^2,
\eal
where $\tilde{\bm x}_t \sim \mc N\pr{\bm x_0, \tilde{\sigma}^2_t\bm I_N}, \tilde{\sigma}^2_t \equiv \tfrac{1 - \abar_t}{\abar_t}$ is defined as a noisy version of $\boldsymbol{x}_0$ within the diffusion process described by~\cref{eq:diffusion_kernel,eq:cumulative_forward} with $T\rightarrow\infty$.
\end{proposition} 

For our proof we use the following intermediate result. 
\begin{lemma}\label{lem:supp_prop1_intermediate}
Let $\bm \sigma \in \R^n_+$, $\bm x \sim p\pr{\bm x} \equiv \mc N\pr{\bm 0, \diag{\bm \sigma^2}}$, $\bm B \in \R^{n\times n}$. Then the following holds:
\bal\label{eq:supp_prop1_lem}
\int \bm x^\transp \bm B \bm x \ p\pr{\bm x}\,d\bm x = \tr{\diag{\bm \sigma^2}\bm B}.
\eal
\end{lemma}
\vspace{-0.5cm}
\begin{proof}[Proof]
We first note that since $\text{Cov}\pr{\bm x} = \diag{\bm \sigma^2}$, the random variables $\bm x_i$ are independent, $p\pr{\bm x} = \prod_{i=1}^n p\pr{\bm x_i}$, and distributed as $\bm x_i \sim p\pr{\bm x_i} \equiv \mc N\pr{0, \bm \sigma^2_i}$. We further note that it holds:
\bal
\bm x^\transp \bm B \bm x = \sum_{i, j = 1}^n \bm x_i \bm x_j \bm B_i^j = \sum_{\substack{i, j = 1\\i\ne j}}^n \bm x_i \bm x_j \bm B_i^j + \sum_{i = 1}^n \bm x^2_i \bm B_i^i.
\eal
Using these results, the derivation of~\cref{eq:supp_prop1_lem} is straightforward:
\bal
&\int \bm x^\transp \bm B \bm x \ p\pr{\bm x}\,d\bm x = \sum_{\substack{i, j = 1\\i \ne j}}^n \Big(\bm B_i^j \int \bm x_i p\pr{\bm x_i}\, d\bm x_i \int \bm x_j p\pr{\bm x_j}\, d\bm x_j \prod_{\substack{k=1\\k\ne i,j}}^n \int p\pr{\bm x_k}\, d\bm x_k\Big) \\\nonumber 
+&\sum_{i = 1}^n \bm B_i^i \int \bm x^2_i p\pr{\bm x_i}\, d\bm x_i \prod_{\substack{k=1\\ k\ne i}}^n \int p\pr{\bm x_k}\, d\bm x_k = \sum_{i = 1}^n \bm B_i^i \bm \sigma^2_i = \tr{\diag{\bm \sigma^2}\bm B},
\eal
since for all $i = \overline{1, n}$, it holds that $\int \bm p\pr{\bm x_i}\, d\bm x_i = 1$, $\int \bm x_i p\pr{\bm x_i}\, d\bm x_i = 0$, $\int \bm x^2_i p\pr{\bm x_i}\, d\bm x_i = \bm \sigma_i^2$.
\end{proof}

We start our proof from computing the left part of the inequality in~\cref{eq:proposition}. Since $\bm y = \bm A \bm x_0 + \bm n$, where $\bm n\sim \mc N\pr{\bm 0, \sigma_{\bm y} \bm I_M}$, we have that: $p\pr{\bm y | \bm x_0} = \mc N\pr{\bm A \bm x_0, \sigma^2_{\bm y}\bm I_M}$. This allows us to write:
\bal\nonumber
&\E_{\bm y | \bm x_0} \lVert \bm x_0 - \bm A^\transp \bm y \rVert_2^2 = \int \lVert \bm x_0 - \bm A^\transp \bm y \rVert_2^2  \ p\pr{\bm y | \bm x_0}\,d\bm y 
= \int \lVert \bm x_0 - \bm A^\transp \bm y \rVert_2^2  \ \frac{\exp\pr{-\frac{\lVert \bm y - \bm A\bm x_0\rVert_2^2}{2\sigma^2_{\bm y}}}}{\pr{\sqrt{2\pi} \sigma_{\bm y}}^M} \,d\bm y \equiv \mc I_1.
\eal
We now apply a change of variables from $\bm y$ to $\bm n = \bm y - \bm A \bm x_0$, for which $d\bm n = d\bm y$. Using the fact that $\frac{1}{\pr{\sqrt{2\pi} \sigma_{\bm y}}^M} \exp\pr{-\frac{\lVert \bm n \rVert_2^2}{2\sigma^2_{\bm y}}} = p\pr{\bm n} = \mc N\pr{\bm 0, \sigma^2_{\bm y}\bm I_M}$, we have that:
\bal\label{eq:supp_prop1_Ey_variables_change}
\mc I_1 = \int \lVert \bm x_0 - \bm A^\transp \bm A\bm x_0 - \bm A^\transp \bm n \rVert_2^2 \ p\pr{\bm n}\,d\bm n.
\eal
Next, we denote $\Delta \bm x_0 \equiv \bm x_0 - \bm A^\transp \bm A \bm x_0 = \pr{\bm I_N - \bm A^\transp \bm A}\bm x_0$ and expand the norm inside the integral to get:
\bal\nonumber 
\mc I_1 = &\Delta \bm x_0^\transp \Delta \bm x_0 \int p\pr{\bm n}\,d\bm n - 2\Delta \bm x_0^\transp \bm A^\transp \int \bm n p\pr{\bm n}\,d\bm n + \int \bm n^\transp \bm A \bm A^\transp \bm n p\pr{\bm n}\,d\bm n.
\eal
Since $p\pr{\bm n} = \mc N\pr{\bm 0, \sigma^2_{\bm y}\bm I_M}$ is a zero-mean Gaussian probability distribution, it holds that $\int p\pr{\bm n}\,d\bm n = 1, \int \bm n p\pr{\bm n}\,d\bm n = 0$, and we get:
\bal
&\mc I_1 = \Delta \bm x_0^\transp \Delta \bm x_0 + \int \bm n^\transp \bm A \bm A^\transp \bm n p\pr{\bm n}\,d\bm n = \Delta \bm x_0^\transp \Delta \bm x_0 + \tr{\sigma^2_{\bm y} \bm A \bm A^\transp} = \Delta \bm x_0^\transp \Delta \bm x_0 + \sigma^2_{\bm y} \lVert \bm A \rVert_F^2,
\eal
where we have used the result of lemma~\ref{lem:supp_prop1_intermediate}. Substituting back the value of $ \Delta \bm x_0$, we end up with:
\bal
\E_{\bm y | \bm x_0} \lVert \bm x_0 - \bm A^\transp \bm y \rVert_2^2 = \lVert\pr{\bm I_N - \bm A^\transp \bm A} \bm x_0\rVert_2^2 + \sigma^2_{\bm y} \lVert \bm A \rVert_F^2.
\eal 

Next, we compute the right part of the inequality in~\cref{eq:proposition}. We first note, that the quantities $\bm y = \bm A \bm x_0 + \bm n$ and $\bm x_t = \sqrt{\abar_t} \bm x_0 + \bm \epsilon_t$ are independent when conditioned on $\bm x_0$, as $\bm n \sim \mc N\pr{\bm 0, \sigma^2_{\bm y}\bm I_M}$ and $\bm \epsilon_t \sim \mc N\pr{\bm 0,  \pr{1-\abar_t}\bm I_N}$ are independent random noise vectors. This allows us to write: $p\pr{\bm y, \bm x_t | \bm x_0} = p\pr{\bm y | \bm x_0}p\pr{\bm x_t | \bm x_0}$. Using these results, the expectation $\E_{\bm y, \bm x_t | \bm x_0} \lVert \bm x_0 - \pr{w \bm A^\transp \bm y + (1-w) \tilde{\bm x}_t}\rVert_2^2$ can be written as:
\bal
&\E_{\bm y, \bm x_t | \bm x_0} \lVert \bm x_0 - \pr{w \bm A^\transp \bm y + (1-w) \tilde{\bm x}_t}\rVert_2^2 \\\nonumber = &\mc I_2 \equiv \int \lVert \bm x_0 - \pr{w \bm A^\transp \bm y + (1-w) \frac{\bm x_t}{\sqrt{\abar_t}}}\rVert_2^2 \ p\pr{\bm y | \bm x_0} p\pr{\bm x_t | \bm x_0}\,d\bm y d\bm x_t.
\eal
Similarly to~\cref{eq:supp_prop1_Ey_variables_change}, we apply the following change of integration variables: $\bm n = \bm y - \bm A\bm x_0,\ d\bm y = d\bm n$ and $\bm \epsilon_t = \bm x_t - \sqrt{\abar_t}\bm x_0,\ d\bm x_t = d\bm \epsilon_t$. We additionally note, that $\bm x_0 - \pr{w \bm A^\transp \bm y + \frac{(1-w)}{\sqrt{\abar_t}} \bm x_t} = \bm x_0 - w\bm A^\transp \bm A \bm x_0 - \pr{1 - w}\bm x_0 - w\bm A^\transp \bm n - \frac{1-w}{\sqrt{\abar_t}}\bm \epsilon_t$, and that $\bm x_0 - w\bm A^\transp \bm A \bm x_0 - \pr{1 - w}\bm x_0 = w\pr{\bm I_N - \bm A^\transp \bm A}\bm x_0 = w\Delta \bm x_0$. As a result, the integral $\mc I_2$ takes the form:
\bal
\mc I_2 = \int \lVert w\Delta \bm x_0 - w\bm A^\transp \bm n - \frac{1-w}{\sqrt{\abar_t}}\bm \epsilon_t\rVert_2^2 \ p\pr{\bm n} p\pr{\bm \epsilon_t}\,d\bm n d\bm \epsilon_t.
\eal
Now we use the augmented variable $$\bm \epsilon = \bbmtx \bm n^\transp& \bm \epsilon_t^\transp\ebmtx^\transp \sim p\pr{\bm n} p\pr{\bm \epsilon_t} = \mc N \pr{\bm 0, \diag{\bbmtx \sigma^2_{\bm y}\bm 1^\transp_M & \pr{1 - \abar_t}\bm 1^\transp_N\ebmtx^\transp}} = p\pr{\bm \epsilon},$$ where with $\bm 1_i$ we denote the vector of dimension $i$ filled with ones. We also denote with $\bm W = \bbmtx w \bm A^T & \frac{1 - w}{\sqrt{\abar_t}} 
\bm I_N\ebmtx$, as in this case, $\bm W \bm \epsilon = w\bm A^\transp \bm n + \frac{1-w}{\sqrt{\abar_t}} \bm \epsilon_t$. With all these modifications, the integral $\mc I_2$ takes the form:
\bal\label{eq:I2}
\mc I_2 = &\int \lVert w\Delta \bm x_0 - \bm W \bm \epsilon \rVert_2^2\ p\pr{\bm \epsilon}\,d\bm\epsilon = w^2 \Delta\bm x_0^\transp \bm x_0 \int p\pr{\bm \epsilon}\,d\bm\epsilon - 2w\Delta \bm x^\transp_0 \bm W \int \bm \epsilon\ p\pr{\bm \epsilon}\,d\bm\epsilon \\\nonumber + & \int \bm \epsilon^\transp \bm W^\transp \bm W \bm \epsilon \ p\pr{\epsilon} \,d\bm\epsilon = w^2 \bm \Delta \bm x^\transp_0 \Delta\bm x_0 + \tr{\diag{\bbmtx \sigma^2_{\bm y}\bm 1^\transp_M & \pr{1 - \abar_t}\bm 1^\transp_N\ebmtx^\transp} \bm W^\transp \bm W}.
\eal
Further, we define $\bm D = \diag{\bbmtx \sigma^2_{\bm y}\bm 1^\transp_M & \pr{1 - \abar_t}\bm 1^\transp_N\ebmtx^\transp}$ and compute the respective trace in~\cref{eq:I2} as:
\bal\nonumber
&\tr{\bm D \bm W^\transp \bm W} = \tr{\bm W\bm D\bm W^\transp} \tr{\pr{\bm W\bm D^{1/2}}\pr{\bm W\bm D^{1/2}}^\transp} \lVert \bm W\bm D^{1/2}\rVert^2_F \\\nonumber 
= \ & w^2\sigma^2_{\bm y}\lVert \bm A\rVert_F^2 + \pr{1-w}^2\frac{1 - \abar_t}{\abar_t}\lVert \bm I_N\rVert_F^2 = w^2\sigma^2_{\bm y}\lVert \bm A\rVert_F^2 + \pr{1-w}^2\frac{1 - \abar_t}{\abar_t}N.
\eal
Based on the above, the integral $\mc I_2$ takes the form:
\bal
\mc I_2 &= w^2 \Delta \bm x^\transp_0\Delta \bm x_0 + w^2\sigma^2_{\bm y}\lVert \bm A\rVert_F^2 + \pr{1-w}^2\frac{1 - \abar_t}{\abar_t} = w^2 \mc I_1 + \pr{1 - w^2}\frac{1-\abar_t}{\abar_t}N.
\eal

As a result of our derivations, the inequality from Proposition~\ref{prop:latent_observation_comparison} takes the form:
\bal\label{eq:supp_prop1_inequality_processed}
& \E_{\bm y | \bm x_0} \lVert \bm x_0 - \bm A^\transp \bm y\rVert_2^2 \le w^2 \E_{\bm y | \bm x_0} \lVert \bm x_0 - \bm A^\transp \bm y\rVert_2^2 + \pr{1-w}^2 \frac{1-\abar_t}{\abar_t}N.
\eal
The quantity $\E_{\bm y | \bm x_0} \lVert \bm x_0 - \bm A^\transp \bm y\rVert_2^2$ has finite value, as it can be upper bounded as follows:
\bal\nonumber
&\E_{\bm y | \bm x_0} \lVert \bm x_0 - \bm A^\transp \bm y \rVert_2^2 = \lVert\pr{\bm I_N - \bm A^\transp \bm A} \bm x_0\rVert_2^2 + \sigma^2_{\bm y} \lVert \bm A \rVert_F^2 \\\nonumber 
\le & \lVert \bm I_N - \bm A^\transp \bm A\rVert_2^2 \lVert \bm x_0\rVert_2^2 + \sigma^2_{\bm y}\min\pr{M, N} \lVert \bm A\rVert_2^2 \\\nonumber 
\le & \pr{\lVert \bm I_N \rVert_2 + \lVert\bm A\rVert_2^2}^2N + \min\pr{M, N} \lVert\bm A\rVert_2^2 \le 4N + \min\pr{M, N}.
\eal
In the above chain of inequalities we have used the submultiplicative property and triangle inequality for $\ell_2$ matrix norms together with the norms equivalence inequality for $\ell_2$ and Frobenius matrix norms, where we also upper-bound the rank of matrix $\bm A$ with its minimal dimension. Additionally, we have used the assumptions of Proposition~\ref{prop:latent_observation_comparison}, specifically $\sigma_{\bm y} \le 1$ and $\lVert \bm A\rVert \le 1$. 

As to the second term of the rhs of~\cref{eq:supp_prop1_inequality_processed}, we note that since $\lim_{t\to \infty}\abar_t = 0$ by design of the diffusion process, it holds that $\lim_{t\to T}\frac{1-\abar_t}{\abar_t} = \lim_{t\to \infty}\frac{1-\abar_t}{\abar_t} = \infty$. This formally translates to the following condition:
\bal
\forall \epsilon > 0 \ \exists \tau\pr{\epsilon} \in \mathbb N: \forall t > \tau \Rightarrow \frac{1-\abar_t}{\abar_t} \ge \epsilon. \label{eq:logical}
\eal
Selection of $\epsilon = \frac{1 + w}{1 - w} \pr{4 + \min\pr{M/N, 1}} \ge \frac{1 + w}{1 - w} \frac{\E_{\bm y | \bm x_0} \lVert \bm x_0 - \bm A^\transp \bm y \rVert_2^2}{N} > 0$ into~\cref{eq:logical} translates it to
\bal
\forall w \in \pr{0, 1} \exists \tau\pr{w} \in \mathbb N: \forall t > \tau \Rightarrow \frac{1-\abar_t}{\abar_t} \ge \frac{\pr{1 - w^2} \E_{\bm y | \bm x_0} \lVert \bm x_0 - \bm A^\transp \bm y \rVert_2^2}{\pr{1 - w}^2 N},
\eal
which concludes the proof given the equivalence of~\cref{eq:supp_prop1_inequality_processed} and~\cref{eq:proposition}.

\section{Proof of Lemma~\ref{lem:single_step}}\label{sec:proposition_single_step}
We prove the correctness of the transition kernel formula by induction on $k$. 
\paragraph{Base case.} First, we focus on the case $k=1$, for which~\cref{eq:single_step} should match the transition probability from~\cref{eq:posterior}. Indeed, from~\cref{eq:single_step_gammas}, it holds:
\bal
&\sum_{i=0}^{k-1} \Gamma_{t-i - 1}^{t-k+1} \gamma_{t - i} = 
\Gamma_{t - 1}^{t} \gamma_{t} = \gamma_{t} = \frac{\sqrt{\abar_{t-1}}\beta_t}{1 - \abar_t},\\
&\Gamma^{t - k + 1}_t = \Gamma^{t}_t = \frac{\sqrt{\alpha_t} \pr{1 - \abar_{t-1}}}{1 - \abar_t},\\
&\sum_{i=0}^{k-1} \pr{\Gamma_{t-i - 1}^{t-k+1}}^2 \sigma^2_{t - i} = \pr{\Gamma_{t - 1}^{t}}^2 \sigma^2_{t} = \sigma^2_t.
\eal
Substituting these results in~\cref{eq:single_step_meanvar} and then in~\cref{eq:single_step} leads us to~\cref{eq:posterior}.
\paragraph{Induction Step.} Let the induction hypothesis from~\cref{eq:single_step} to be valid for some $k$, such that $t - k \in \left(0, T\right]$. Then, we need to show that \cref{eq:single_step} is also valid for $k + 1$. 

We note, that from~\cref{eq:posterior} we have $\bm x_{t - k - 1} = \bm \mu_{t-k}\pr{\bm x_{t-k}, \bm x_0} + \sigma_{t - k}\bm \epsilon_1$, where $\bm \epsilon_1 \sim \mc N\pr{\bm 0, \bm I_N}$. At the same time, since~\cref{eq:single_step} is valid for $k$ by the induction hypothesis, we have that $\bm x_{t - k} = \bm \mu_{t,k}\pr{\bm x_t, \bm x_0} + \sigma_{t,k}\bm \epsilon_2$, where $\bm \epsilon_2 \sim \mc N\pr{\bm 0, \bm I_N}$. Combining these two equations with ~\cref{eq:posterior,eq:single_step,eq:single_step_meanvar}, we get:
\bal\label{eq:supp_lem2_induction_step_main}
&\bm x_{t - k - 1} = \frac{\sqrt{\abar_{t-k-1}}\beta_{t-k}}{1 - \abar_{t-k}} \bm x_0 + \frac{\sqrt{\alpha_{t-k}}\pr{1 - \abar_{t-k-1}}}{1 - \abar_{t-k}} \bm x_{t-k} + \sigma_{t-k} \bm \epsilon_1 \\\nonumber
= \ &\bm x_0 \left(\frac{\sqrt{\alpha_{t-k}}\pr{1 - \abar_{t-k-1}}}{1 - \abar_{t-k}}\sum_{i=0}^{k-1} \Gamma_{t-i - 1}^{t-k+1} \gamma_{t-i} + \frac{\sqrt{\abar_{t-k-1}}\beta_{t-k}}{1 - \abar_{t-k}}\right) + \bm x_t \frac{\sqrt{\alpha_{t-k}}\pr{1 - \abar_{t-k-1}}}{1 - \abar_{t-k}} \Gamma_t^{t-k+1} \\\nonumber 
+ \ &\sqrt{\frac{\alpha_{t-k}\pr{1 - \abar_{t-k-1}}^2}{\pr{1 - \abar_{t-k}}^2} \sum_{i=0}^{k-1} \pr{\Gamma_{t-i-1}^{t-k+1}}^2 \sigma_{t - i}^2}\bm \epsilon_2 + \sigma_{t-k}\bm \epsilon_1.
\eal

To move further, we first prove that $\forall i,j\in \mathbb{N}^+, i\ge j$, $\Gamma_i^j$ from~\cref{eq:single_step_gammas} can be decomposed as: $\Gamma_i^j = \Gamma_j^j \Gamma_i^{j+1}$. This is easy to show by direct substitution, that is:
\bal\label{eq:supp_lem2_gamma_cap_decomposition}
&\Gamma_j^j \Gamma_i^{j+1} = \sqrt{\alpha_j} \frac{1 - \abar_{j-1}}{1 - \abar_j} \sqrt{\prod_{n=j+1}^i\alpha_n} \frac{1 - \abar_{j}}{1 - \abar_i} \sqrt{\alpha_j \prod_{n=j+1}^i\alpha_n} \frac{1 - \abar_{j-1}}{1 - \abar_i} =\sqrt{\prod_{n=j}^i\alpha_n} \frac{1 - \abar_{j-1}}{1 - \abar_i} = \Gamma_i^j.
\eal
We additionally note that it holds:
\bal\label{eq:supp_lem2_multipier_via_gamma_cap}
&\frac{\sqrt{\alpha_{t-k}}\pr{1 - \abar_{t-k-1}}}{1 - \abar_{t-k}} = \Gamma_{t - k}^{t - k},\\\label{eq:supp_lem2_multipier_via_gamma_small}
&\frac{\sqrt{\abar_{t-k-1}}\beta_{t-k}}{1 - \abar_{t-k}} = \gamma_{t - k} = \Gamma_{t - k - 1}^{t - k} \gamma_{t - k}\\\label{eq:supp_lem2_sigma_with_gamma_cap}
&\sigma^2_{t-k} = \pr{\Gamma_{t - k - 1}^{t - k}}^2 \sigma^2_{t-k},
\eal
as $\Gamma_{t - k - 1}^{t - k} = 1$, since $t - k - 1 < t - k$.

First, we compute the multiplier of $\bm x_t$ in~\cref{eq:supp_lem2_induction_step_main}. Since $t \ge t-k+1 \ \ \forall k \in \mathbb{N}^+$, we can use~\cref{eq:supp_lem2_gamma_cap_decomposition} to derive the following result:
\bal\label{eq:supp_lem2_xt_mult_final}
&\frac{\sqrt{\alpha_{t-k}}\pr{1 - \abar_{t-k-1}}}{1 - \abar_{t-k}} \Gamma_t^{t-k+1} = \Gamma_{t - k}^{t - k}\Gamma_t^{t-k+1} = \Gamma_{t}^{t-k},
\eal
where we have additionally utilized the result of~\cref{eq:supp_lem2_multipier_via_gamma_cap}.
Next, we simplify the multiplier of $\bm x_0$ from~\cref{eq:supp_lem2_induction_step_main}. We divide the sum inside this multiplier into two parts: 
\bal
&\sum_{i=0}^{k-1} \Gamma_{t-i - 1}^{t-k+1} \gamma_{t-i} = \sum_{i=0}^{k-2} \Gamma_{t-i - 1}^{t-k+1} \gamma_{t-i} + \Gamma_{t-k}^{t-k+1} \gamma_{t-k+1} = \label{eq:supp_lem2_sum_separation}
\sum_{i=0}^{k-2} \Gamma_{t-i - 1}^{t-k+1} \gamma_{t-i} + \gamma_{t-k+1},
\eal
where to separate the $\gamma_{t-k+1}$ term we have used the lower case of $\Gamma^j_i$ from~\cref{eq:single_step_gammas}. Here we additionally note, that if $k = 1$, then the remaining sum $\sum_{i=0}^{k-2} \Gamma_{t-i - 1}^{t-k+1}$ becomes zero. If $k > 1$, then for all the terms of this sum it holds $t - i - 1 \ge t - k + 1$. Combining this with the results of~\cref{eq:supp_lem2_gamma_cap_decomposition,eq:supp_lem2_multipier_via_gamma_cap,eq:supp_lem2_multipier_via_gamma_small}, we can compute the multiplier of $\bm x_0$ from~\cref{eq:supp_lem2_induction_step_main} as:
\bal\nonumber
&\frac{\sqrt{\alpha_{t-k}}\pr{1 - \abar_{t-k-1}}}{1 - \abar_{t-k}}\sum_{i=0}^{k-1} \Gamma_{t-i - 1}^{t-k+1} \gamma_{t-i} + \frac{\sqrt{\abar_{t-k-1}}\beta_{t-k}}{1 - \abar_{t-k}} = \sum_{i=0}^{k-2} \Gamma_{t-k}^{t-k}\Gamma_{t-i - 1}^{t-k+1}\gamma_{t-i} + \Gamma_{t-k}^{t-k} \gamma_{t - k + 1} \\\label{eq:supp_lem2_x0_mult_final} + 
&\Gamma_{t-k-1}^{t-k}\gamma_{t-k} = \sum_{i=0}^{k-2} \Gamma_{t-i - 1}^{t-k} \gamma_{t-i} + \br{\Gamma_{t-i - 1}^{t-k} \gamma_{t-i}}_{i=k-1} + \br{\Gamma_{t-i - 1}^{t-k} \gamma_{t-i}}_{i=k} =\sum_{i=0}^{k} \Gamma_{t-i - 1}^{t-k} \gamma_{t-i}.
\eal

To simplify the remaining part of~\cref{eq:supp_lem2_induction_step_main}, which involves the terms with the Gaussian noise vectors $\bm \epsilon_1$ and $\bm \epsilon_2$, we utilize the fact that $\forall a, b \ge 0, \ \ \bm \epsilon_1, \bm \epsilon_2 \sim \mc N\pr{\bm 0, \bm I_N}$ it holds: $\sqrt{a} \bm \epsilon_1 + \sqrt{b} \bm \epsilon_2 \sim \mc N\pr{\bm 0, \pr{a + b}\bm I_N}$. Based on this, we can write:
\bal
&\sqrt{\frac{\alpha_{t-k}\pr{1 - \abar_{t-k-1}}^2}{\pr{1 - \abar_{t-k}}^2} \sum_{i=0}^{k-1} \pr{\Gamma_{t-i-1}^{t-k+1}}^2 \sigma_{t - i}^2}\bm \epsilon_2 + \sigma_{t-k}\bm \epsilon_1 \sim \mc N \pr{\bm 0, \hat{\sigma}^2_{t, k+1} \bm I},
\eal
where, if we additionally use the results of~\cref{eq:supp_lem2_multipier_via_gamma_cap,eq:supp_lem2_sigma_with_gamma_cap}, we have that:
\bal
\hat{\sigma}^2_{t, k+1} = \pr{\Gamma_{t-k}^{t-k}}^2 \sum_{i=0}^{k-1} \pr{\Gamma_{t-i-1}^{t-k+1}}^2 \sigma_{t - i}^2 + \pr{\Gamma^{t-k}_{t-k-1}}^2 \sigma_{t-k}^2.
\eal
Now, we use the same strategy as in~\cref{eq:supp_lem2_sum_separation} and~\cref{eq:supp_lem2_x0_mult_final} to simplify this expression: 
\bal\label{eq:supp_lem2_sigma_sum_final}
&\hat\sigma^2_{t, k+1} = \sum_{i=0}^{k-2} \pr{\Gamma_{t-k}^{t-k}\Gamma_{t-i-1}^{t-k+1}}^2 \sigma_{t - i}^2 + \pr{\Gamma_{t-k}^{t-k}}^2 \sigma^2_{t-k+1} + \pr{\Gamma^{t-k}_{t-k-1}}^2 \sigma_{t-k}^2 = \sum_{i=0}^{k} \pr{\Gamma_{t-i-1}^{t-k}}^2 \sigma_{t - i}^2.
\eal

\noindent From~\cref{eq:supp_lem2_xt_mult_final,eq:supp_lem2_x0_mult_final,eq:supp_lem2_sigma_sum_final}, we get:
\bal
\bm x_{t - k - 1} \sim \mc N\pr{\hat{\bm\mu}_{t, k+1}\pr{\bm x_t,\bm x_0}, \hat{\sigma}^2_{t,k+1}\bm I_N},
\eal
where
\bal
&\hat{\bm\mu}_{t, k+1}\pr{\bm x_t,\bm x_0} = \sum_{i=0}^{k} \Gamma_{t-i - 1}^{t-k} \gamma_{t-i} \bm x_0 + \Gamma_{t}^{t-k} \bm x_t,\\
&\hat\sigma^2_{t, k+1} = \sum_{i=0}^{k} \pr{\Gamma_{t-i-1}^{t-k}}^2 \sigma_{t - i}^2.
\eal
By direct substitution of $k=k+1$ into~\cref{eq:single_step_meanvar} it is easy to show that
\bal\nonumber
&\hat{\bm\mu}_{t, k+1} \pr{\bm x_t,\bm x_0} = \bm\mu_{t, k+1} \pr{\bm x_t,\bm x_0},\\\nonumber
&\hat\sigma^2_{t, k+1} = \sigma^2_{t, k+1},
\eal
which implies that the induction hypothesis from~\cref{eq:single_step} is valid for $k+1$. This completes the proof of the induction step and combined with with the base case it proves by induction the validity of~\cref{eq:single_step} for every feasible $k$.

\section{Comparative Analysis: Proposed Accelerated Sampling and Prior work} 
\label{sec:appendix_acceleration_difference}
\subsection{Conceptual Difference}
Using our notation, both~\cite{Chung2022come} and~\cite{Meng2022sdedit} propose to start the reverse process from a timestep $\tau$ and a noisy version $\boldsymbol{x}_\tau$ of the initial estimate of $\boldsymbol{x}_0$, which we denote by $\mathbb{E}\left[\boldsymbol{x}_0| \boldsymbol{y}\right]$. The main conceptual difference of our approach is that in these cases $\boldsymbol{x}_\tau$ is obtained using the forward diffusion process, while in our case we end up in $\boldsymbol{x}_\tau$ using the reverse process. The initial motivation for our proposed approach is also different. In particular, while we motivate our procedure from a probabilistic viewpoint and propose to approximate the conditional score function as a composition of three functions, the authors in ~\cite{Chung2022come} base their strategy on the contrastive property of reverse SDEs, while the authors in ~\cite{Meng2022sdedit} use the re-projection of unrealistic images to the manifold of natural images in the noisy latent space.

\subsection{Technical Difference}
Given that in our work we consider the standard DDPM realization of diffusion process (VP-SDE), we will explain the existing differences under this scenario. The authors of~\cite{Chung2022come} and~\cite{Meng2022sdedit} propose to parameterize $\boldsymbol{x}_\tau$ as $$\boldsymbol{x}_\tau=\sqrt{\bar{\alpha}_\tau}\mathbb{E}\left[\boldsymbol{x}_0|\boldsymbol{y}\right]+\sqrt{1-\bar{\alpha}_\tau} \boldsymbol{z},\ \ \boldsymbol{z}\sim\mathcal{N}\left(\boldsymbol{0},\boldsymbol{I}\right).$$ In contrast, in our case by using Eq.(10) we adopt the following parametrization: $$\boldsymbol{x}_\tau=\sum_{i=0}^{T-\tau-1}\Gamma_{T-i-1}^{\tau+1}\gamma_{T-i}\mathbb{E}\left[\boldsymbol{x}_0|\boldsymbol{y}\right]+\Gamma_T^{\tau+1}\boldsymbol{x}_T + \sqrt{\sum_{i=0}^{T-\tau-1}\left(\Gamma_{T-i-1}^{\tau+1}\right)^2\sigma_{T - i}^2}\boldsymbol{z},\ $$ where $\boldsymbol{z},\boldsymbol{x_T}\sim\mathcal{N}\left(\boldsymbol{0},\boldsymbol{I}\right).$

Thus, our parametrization is more general and it is possible to show by induction, that under certain conditions it leads to the exact same $\boldsymbol{x}_\tau$ as in~\cite{Chung2022come} and~\cite{Meng2022sdedit}.

Finally, to experimentally demonstrate that our approach exhibits certain benefits compared to the ones described in~\cite{Chung2022come} and~\cite{Meng2022sdedit}, we conducted additional comparisons for the SISR problem between the different sampling strategies (see Table~\ref{table:acceleration_ablation}). From these results it is clear that our proposed strategy works better in practice and leads to superior results both in terms of fidelity and perceptual quality.

\begin{table}[!h]
\caption{Comparison of the proposed acceleration scheme and prior works ~\cite{Chung2022come, Meng2022sdedit} for the SISR task.}
\label{table:acceleration_ablation}
\tts
\centering
\begin{tabular}{l|llll|l}
\hline
Acceleration Strategy  & PSNR $\uparrow$ & SSIM $\uparrow$  & LPIPS $\downarrow$ & TOPIQ$_{\Delta}$ $\downarrow$ & NFE $\downarrow$ \\ \hline
Ours              & \bf{28.12} & \bf{0.793} & \bf{0.140} & \bf{0.002} & 51  \\ 
~\cite{Chung2022come} and~\cite{Meng2022sdedit} & 28.05 & 0.783 & 0.142 & 0.016 & 51  \\ \hline
\end{tabular}
\end{table}

\section{Computational Cost Analysis} \label{sec:appendix_complexity}
In this section, we calculate the computational cost for each diffusion-based method in terms of TFLOPs. The input image size for all competing approaches is 720p (1280 x 720).
\begin{table}[h!]
\caption{Computational cost of the proposed and existing diffusion-based methods for the Dynamic Scene Deblurring task with 720p input resolution.}
\label{table:computational_complexity}
\centering
\begin{tabular}{l|l|l}
\hline
Method          &  TFLOP (equation) & TFLOP Total $\downarrow$\\ \hline
DvSR~\cite{whang2022deblurring}  & 1.2$\times$NFE + 4.8    &   604.8  \\ 
icDPM~\cite{ren2023multiscale} & 4.8$\times$NFE + 5.2    &  2405.2 \\ 
InDI~\cite{delbracio2023inversion}  & 4.8$\times$NFE          &    48.0 \\ 
Ours & 4.3$\times$NFE + 1.9     &    \textbf{23.4}  \\ \hline
\end{tabular}
\end{table}

The proper way to interpret equations in Table~\ref{table:computational_complexity} is as follows: $\textrm{TFLOP}_{\textrm{total}} = x \times N + y$, where $x$ is the TFLOP complexity for a single backbone pass within the diffusion process, $N$ is the total number of neural function evaluations (NFEs) per sampling process, and 
 is the complexity of sub-modules that have to be run once per image (e.g. Image Restoration network in our method, pre-processing net for icDPM~\cite{ren2023multiscale} and DvSR~\cite{whang2022deblurring}). Based on these results, we observe that the computational cost of our method is significantly lower compared to our diffusion-based competitors.

\section{Proposed one-step acceleration with/without DDIM} \label{sec:appendix_ddim_combo}
Utilizing our one-step acceleration, we bypass steps from \( t = T \) to \( t = \tau \), allowing us to either straightforwardly execute the remaining \( t = \tau \) steps or apply any existing acceleration strategies. To demonstrate that our one-step acceleration is complementary to existing accelerated sampling strategies, we combined the DDIM~\cite{song2021denoising} acceleration technique with our one-step acceleration for the single-image super-resolution (SISR) task (refer to Table~\ref{table:ddim_acceleration}). We notice that no significant quantitative/qualitative difference after applying this acceleration technique has been observed. 

\begin{table}[h!]
\caption{Results for the proposed one-step acceleration with/without DDIM acceleration tested on SISR task}
\label{table:ddim_acceleration}
\centering
\begin{tabular}{l|llll|l}
\hline
Method       & PSNR $\uparrow$  & SSIM $\uparrow$  & LPIPS $\downarrow$ & TOPIQ$_{\Delta} $ $\downarrow$ & NFE $\downarrow$ \\ \hline
Ours (with DDIM) & 28.12 & 0.793 & 0.140 & \textbf{0.002} & 51  \\
Ours w/o DDIM & \textbf{28.16} & \textbf{0.794} & \textbf{0.139} & 0.007 & 251 \\ \hline
\end{tabular}
\end{table}

\section{Training Procedure} \label{sec:supp_training}
\paragraph{Denoising Module.} We use two denoising architectures, namely MIRNet-S and UDP~\cite{brooks2019unprocessing} for our main experiments and ablation studies. MIRNet-S (15.95M parameters) is a lighter version of the original MIRNet (31.78M parameters)~\cite{zamir2020learning}, where MSRB is decreased from 2 to 1. In contrast, UDP (11.78M parameters)~\cite{brooks2019unprocessing} architecture is not modified and used as it is. Both models are trained for the Gaussian denoising task in the sRGB domain with input noise level $\tilde{\sigma} \in [0, 244.3]$. More specifically, for each element of batch the noise standard deviation is selected randomly using uniform sampling within this range. The batch size, training crop size, and initial learning rate are set to 8 (in total), $192\times192$, and $2\cdot10^{-4}$, respectively. Overall, all denoisers are trained for 1M iterations on 8 Ascend 910 AI accelerators 
using the Adam~\cite{kingma2014adam} optimizer with default parameters and a decaying learning rate scheduler: $lr_s = lr_0*\gamma^{\floormod{s/2000}}$, where $\gamma=0.999$. We employ an MSE loss to train the denoising model and concatenate the noise level with the noisy image as an input to the denoiser same as ~\cite{brooks2019unprocessing}. 
\paragraph{IR Module.} As mentioned in the main manuscript, we utilize existing models with publicly available pretrained parameters. For the tasks of burst JDD-SR, dynamic scene deblurring, and SISR we have used BSRT-Small~\cite{luo2022bsrt}, FFTFormer~\cite{kong2023efficient}, and  SwinIR~\cite{liang2021swinir}, respectively. A description of each one of these IR networks, including their number of trainable parameters, is provided in Table~\ref{table:parameters}.
\paragraph{Fusion Module.} For each IR task under study, we follow the same protocol. We train the fusion module for $300K$ iterations with batch size of 128 (in total), and crop size of $256 \times 256$. The training takes place on 8 Ascend 910 AI accelerators, 
while the optimizer of choice is Adam~\cite{kingma2014adam} with default parameters and a learning rate scheduler $lr_s = lr_0*\gamma^{\floormod{s/1000}}$, where $\gamma=0.99$. For each one of the studied IR tasks we train our Fusion module on a dedicated dataset. Specifically, we use 
the ZurichRaw2RGB~\cite{ignatov2020replacing} dataset for burst JDD-SR, GoPro~\cite{nah2017deep} for dynamic scene deblurring, and DIV2K~\cite{agustsson2017ntire} for SISR. The selection of these specific datasets is motivated by the fact that they are widely used by all the competing methods for network training related to the IR tasks of interest. The detailed description of training data we used for each problem is provided below.

\begin{itemize}
\item \textbf{JDD-SR.} Following the same protocol as in~\cite{dbsr21, lecouat2021lucas, luo2021ebsr, deeprep21, dudhane2022burst, luo2022bsrt}, we generate~46K burst sets from the training set of ZurichRAW2RGB. Each burst set, which contains 14 low-resolution images in the raw domain, is inferenced by BSRT-Small~\cite{luo2022bsrt} and post-processed to produce an image in the sRGB domain. Then, these predictions are used as input to train \textbf{only} our Fusion module, while Denoising and IR networks are frozen.

\item \textbf{Dynamic Scene Deblurring.} We follow the standard protocol for this problem and use the GoPro dataset for training. Specifically, we use 3214 pairs of clean and blurry $1280 \times 720$ images, out of which we have excluded the 1111 pairs reserved for evaluation purposes. In order to provide a fair comparison, we follow exactly the same setup as in~\cite{kupyn2018deblurgan,zhang2020deblurring,zamir2022restormer,kong2023efficient} and train \textbf{only} the fusion module using the provided GoPro training data. 

\item \textbf{SISR} We employ the well-known DIV2K~\cite{agustsson2017ntire} dataset for the SISR task. This dataset contains a set of 800 images of 2K resolution, which we used for training our Fusion module. Following the standard protocol, we use the additionally provided 100 2K images for evaluation. 
\end{itemize}

\label{sec:supp_mat_fuser}
\begin{figure}[!h]
    \centering
    \includegraphics[width=0.7\linewidth]{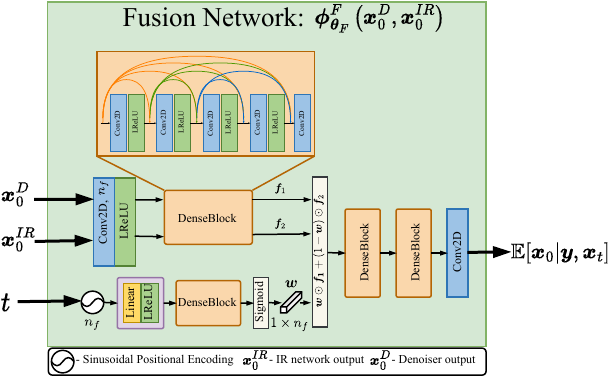}
    \caption{Detailed visualization of the proposed Fusion module.}
    \label{fig:fuser_detailed}
\end{figure}
\section{Fusion Module}\label{sec:appendix_fusion_module} 
The main goal of our proposed Fusion module is to predict the conditional expectation $\E\br{\bm x_0 | \bm y, \bm x_t}$ given the estimates of $\E\br{\bm x_0 | \bm x_t}$ and $\E\br{\bm x_0 | \bm y}$, which are produced by the denoising and IR modules, respectively. To do so, our fusion network accepts as inputs the image estimates $\bm x_0^{D}, \bm x_0^{IR}$ and a timestep $t$. Its exact architecture is  depicted in Figure~\ref{fig:fuser_detailed} and it consists of two branches. The upper branch operates on $\bm x_0^{D}$, $\bm x_0^{IR}$ and produces the corresponding features $\bm f_1$, $\bm f_2$ using a single dense block~\cite{huang2017densely} without a normalization layer. The lower branch encodes the timestep $t$ into a vector of weights $\bm w \in \pr{0, 1}^{n_f}$ using the sinusoidal positional encoding~\cite{vaswani2017attention}, followed by a two layer MLP, and a sigmoid function as the final activation. Then, we perform a weighted summation of $\bm f_1$, $\bm f_2$ in the feature space. Finally, two consequent dense blocks~\cite{huang2017densely}, with $n_f$ channels each, followed by a convolution layer produce the final estimate of $\E\br{\bm x_0 | \bm y, \bm x_t}$. Our proposed architecture has only 0.73M learnable parameters, which is significantly lower (7-22 times) compared to the Denoising and IR modules. As a result, the Fusion module requires less training time/resources and can be trained only on a small amount of problem-specific training data.

\begin{table}[!t]
\caption{Number of parameters and source code of methods used as an IR module in our framework.}
\centering
\label{table:parameters}
\begin{tabular}{l|c|c}
\hline 
IR Modules   & Parameters $\downarrow$ & Code \& Weights \\ \hline
BSRT-Small   & 4.92M  & \href{https://github.com/Algolzw/BSRT}{link}         \\
FFTFormer    & 16.56M & \href{https://github.com/kkkls/FFTformer}{link}      \\
SwinIR       & 11.85M & \href{https://github.com/JingyunLiang/SwinIR}{link}   \\ \hline
\end{tabular}
\end{table}

\section{Fusion Module Robustness}
\label{supp_ablation:fusion_robustness}
In order to assess the robustness/generalization ability of our method, we have conducted additional experiments where we evaluate the reconstruction quality achieved when the Fusion network, which was originally trained on the MIRNet-S and SwinIR pair, is combined with the following pairs of denosing and IR networks:
\begin{itemize}
\item The Fusion module is combined with the same pair of denoising and IR networks as those used during its training.
\item The Fusion module is combined with a different IR network and the same denoising network as the one used during its training.
\item The Fusion module is combined with a different denoising network and the same IR network as the one used during its training.
\item The Fusion module is combined with different denoising and IR networks than the ones used during its training.
\end{itemize}

\begin{table}[!h]
\centering
\caption{The performance of Fusion module for different train/test pair scenarios for 4x SR task.}

\label{table:fusion_test_train_ablation}
\begin{tabular}{l|ll|cccc}
\hline
$\textnumero$ & Trained Pair           & Tested Pair       & PSNR\tts{$\uparrow$} & SSIM\tts{$\uparrow$}  & LPIPS\tts{$\downarrow$} & TOPIQ$_{\Delta}$\tts{$\downarrow$} \\ \hline
-             & \multicolumn{2}{c|}{Target}                & $\infty$   & 1     & 0     & 0     \\ \hline
1             & MIRNet-S + SwinIR      & MIRNet-S + SwinIR & 28.12 & 0.793 & 0.140 & 0.002 \\
2             & MIRNet-S + SwinIR      & MIRNet-S + RRDB   & 28.20 & 0.795 & 0.144 & 0.023 \\
3             & MIRNet-S + SwinIR      & UDP + SwinIR      & 28.47 & 0.808 & 0.177 & 0.017 \\
4             & MIRNet-S + SwinIR      & UDP + RRDB        & 28.46 & 0.807 & 0.183 & 0.025 \\ \hline
\end{tabular}
\end{table}

From these experiments as shown in the table above, we observe that changing the denoising network to a less powerful one leads to a noticeable drop in terms of perceptual quality (~30\% in LPIPS) and slightly blurrier results, which corresponds to a PSNR increase by 0.2dB. Conversely, altering the IR network has a minimal impact on both the reconstruction and perceptual metrics. In summary, under the particular IR task, the Fusion network evaluated with a different pair of denoising and IR modules demonstrates a good generalization ability and a robust behavior.

\section{Perception-Fidelity Trade-off}
\label{supp_ablation:tradeoff_gopro}
We have evaluated our method on the GoPro dataset (motion deblurring) for different $\tau$, when the denoising and Fusion modules are activated. From Table~\ref{table:ablation_tau_gopro}, we observe that the perception-fidelity trade-off follows a similar trend to the one for the SISR 4x task. The main difference is that in this particular case the necessary reverse diffusion steps are less than in SISR.
\begin{table}[!h]
\centering
\caption{Perception-Distortion Trade-off on GoPro validation for dynamic scene deblurring task.}
\label{table:ablation_tau_gopro}
\begin{tabular}{l|cccccc}
\hline
\textbf{$\tau$} & \textbf{0} & \textbf{1} & \textbf{5} & \textbf{10} & \textbf{20} & \textbf{25} \\ \hline
PSNR, dB        & 34.21      & 34.02      & 33.72      & 33.52       & 33.23       & 33.14       \\
LPIPS           & 0.071      & 0.057      & 0.053      & 0.052       & 0.052       & 0.052       \\ \hline
\end{tabular}
\end{table}

\section{Information about Competing methods.}
Below, we provide links to the code implementation and trained weights for all baselines used for comparison.

\begin{table}[h!]
\centering
\caption{Code and model weights of the competing methods for burst JDD-SR task.}
\label{table:links_to_burst_methods}
\begin{tabular}{l|l|l}
Method  & Link to Code                               & Link to Weights \\ \hline
DBSR    & \href{https://github.com/goutamgmb/deep-burst-sr}{link} & \href{https://drive.google.com/file/d/1bdtz\_gr\_m9MnypVqDigH6H1xoae82lwY/view}{link} \\
DeepRep & \href{https://github.com/goutamgmb/deep-rep}{link}      & \href{https://drive.google.com/file/d/1GZ0S4BKaZbgiy9WmoYH4kAsXY7zDR\_ks/view}{link} \\
EBSR    & \href{https://github.com/Algolzw/EBSR}{link}            & \href{https://drive.google.com/file/d/1Zz21YwNtiKZCjerrZsdvcWyubqTJBwaD/view}{link} \\
BIPNet  & \href{https://github.com/akshaydudhane16/BIPNet}{link}  & \href{https://mbzuaiac-my.sharepoint.com/personal/akshay\_dudhane\_mbzuai\_ac\_ae/\_layouts/15/onedrive.aspx?id=\%2Fpersonal\%2Fakshay\%5Fdudhane\%5Fmbzuai\%5Fac\%5Fae\%2FDocuments\%2FCVPR\%2D22\%2FBurst\%20Super\%2Dresolution\%2FTrained\%5Fmodels\%2FTrained\%5Fmodels\%2FReal\%2FBIPNet\%2Epth\&parent=\%2Fpersonal\%2Fakshay\%5Fdudhane\%5Fmbzuai\%5Fac\%5Fae\%2FDocuments\%2FCVPR\%2D22\%2FBurst\%20Super\%2Dresolution\%2FTrained\%5Fmodels\%2FTrained\%5Fmodels\%2FReal\&ga=1}{link} \\
BSRT    & \href{https://github.com/Algolzw/BSRT}{link}            & \href{https://drive.google.com/file/d/1Bv1ZwoE3s8trhG--wjB0Yt6WJIQPpvsn/view}{link} \\ \hline
\end{tabular}
\end{table}

\begin{table}[h!]
\centering
\caption{Code and model weights of the competing methods for dynamic scene deblurring task.}
\label{table:links_to_deblur_methods}
\begin{tabular}{l|l|l}
Method  & Link to Code                               & Link to Weights \\ \hline
HINet       & \href{https://github.com/megvii-model/HINet}{link}     & \href{https://drive.google.com/file/d/1dw8PKVkLfISzNtUu3gqGh83NBO83ZQ5n/view}{link}            \\
MPRNet      & \href{https://github.com/swz30/MPRNet}{link}           & \href{https://drive.google.com/file/d/1QwQUVbk6YVOJViCsOKYNykCsdJSVGRtb/view}{link}            \\
MIMO-UNet+  & \href{https://github.com/chosj95/MIMO-UNet}{link}      & \href{https://drive.google.com/file/d/166sufeHcdDTgXHNbCRzTC4T6DzuflB5m/view}{link}            \\
NAFNet      & \href{https://github.com/megvii-research/NAFNet}{link} & \href{https://drive.google.com/file/d/1S0PVRbyTakYY9a82kujgZLbMihfNBLfC/view}{link}            \\
Restormer   & \href{https://github.com/swz30/Restormer}{link}        & \href{https://drive.google.com/drive/folders/1czMyfRTQDX3j3ErByYeZ1PM4GVLbJeGK}{link}          \\
FFTFormer   & \href{https://github.com/kkkls/FFTformer}{link}        & \href{https://github.com/kkkls/FFTformer/blob/main/pretrain\_model/fftformer\_GoPro.pth}{link} \\
DeblurGANv2 & \href{https://github.com/VITA-Group/DeblurGANv2}{link} & \href{https://github.com/VITA-Group/DeblurGANv2}{link} \\ \hline
\end{tabular}
\end{table}

\begin{table}[h!]
\centering
\caption{Code and model weights of the competing methods for SISR task.}
\label{table:links_to_sisr_methods}
\begin{tabular}{l|l|l}
Method  & Link to Code                               & Link to Weights \\ \hline
SRResNet       & \href{https://github.com/twtygqyy/pytorch-SRResNet}{link}     & \href{https://github.com/twtygqyy/pytorch-SRResNet/blob/master/model/model_srresnet.pth}{link}            \\
RRDB      & \href{https://github.com/xinntao/ESRGAN}{link}           & \href{https://drive.google.com/file/d/1pJ_T-V1dpb1ewoEra1TGSWl5e6H7M4NN/view?usp=share_link}{link}            \\
SwinIR  & \href{https://github.com/JingyunLiang/SwinIR}{link}      & \href{https://github.com/JingyunLiang/SwinIR/releases/download/v0.0/001_classicalSR_DF2K_s64w8_SwinIR-M_x4.pth}{link}            \\
LIIF      & \href{https://github.com/yinboc/liif}{link} & \href{https://drive.google.com/file/d/1xaAx6lBVVw_PJ3YVp02h3k4HuOAXcUkt/view?usp=sharing}{link}            \\
HAT   & \href{https://github.com/XPixelGroup/HAT}{link}        & \href{https://drive.google.com/file/d/1pdhaO1fJq3tgSqDIbymdDiGxu4S0nqVq/view?usp=share_link}{link}          \\
ESRGAN   & \href{https://github.com/xinntao/ESRGAN}{link}        & \href{https://drive.google.com/file/d/1TPrz5QKd8DHHt1k8SRtm6tMiPjz_Qene/view?usp=share_link}{link} \\
HCFlow & \href{https://github.com/JingyunLiang/HCFlow}{link} & \href{https://github.com/JingyunLiang/HCFlow/releases/download/v0.0/SR_DF2K_X4_HCFlow.pth}{link} \\ 
SwinIR-GAN & \href{https://github.com/JingyunLiang/SwinIR}{link} & \href{https://github.com/JingyunLiang/SwinIR/releases/download/v0.0/003_realSR_BSRGAN_DFOWMFC_s64w8_SwinIR-L_x4_GAN.pth}{link} \\ 
LDM & \href{https://github.com/CompVis/latent-diffusion}{link} & \href{https://ommer-lab.com/files/latent-diffusion/sr_bsr.zip}{link} \\ 
SRDiff & \href{https://github.com/LeiaLi/SRDiff}{link} & \href{https://github.com/LeiaLi/SRDiff/releases/download/v1.0.0/srdiff_pretrained_div2k.zip}{link} \\  
IDM & \href{https://github.com/Ree1s/IDM}{link} & \href{https://drive.google.com/file/d/1tWUVyMwV0eHPPDdz-RYIDxJ_QwMUhTHM/view?usp=share_link}{link} \\ \hline
\end{tabular}
\end{table}

\section{Additional Results} \label{sec:supp_addition_imgs}
In the section, we provide additional visual comparison of the proposed method with existing SOTA approaches for all IR tasks under study. 
\begin{figure*}[!h]
\centering
\setlength\extrarowheight{-10pt}
\begin{tabular}{@{\hskip 0.05cm} c @{\hskip 0.05cm} c @{\hskip 0.05cm} c @{\hskip 0.05cm} c @{\hskip 0.05cm} c @{\hskip 0.05cm}}
  \multirow{3}{*}[1.845cm]{\includegraphics[width=.293\linewidth]{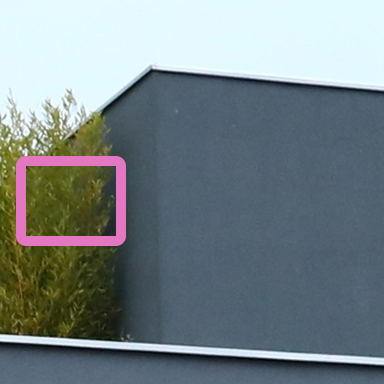}} &
 \includegraphics[width=.170\linewidth]{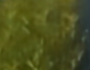} & 
 \includegraphics[width=.170\linewidth]{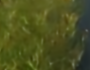} & 
 \includegraphics[width=.170\linewidth]{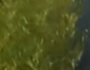} &
 \includegraphics[width=.170\linewidth]{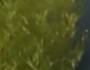} \\
 & \scriptsize{DBSR: 0.144} & \scriptsize{DeepRep: 0.084} & \scriptsize{BSRT-Small: 0.048} & \scriptsize{BSRT-Large: 0.060} \\
 
 & \includegraphics[width=.170\linewidth]{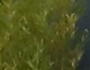} & 
 \includegraphics[width=.170\linewidth]{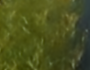} & 
 \includegraphics[width=.170\linewidth]{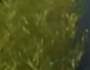} &
 \includegraphics[width=.170\linewidth]{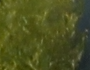} \\
  & \scriptsize{Target} & \scriptsize{BIPNet: 0.069} & \scriptsize{EBSR: 0.054} & \scriptsize{Ours: \textbf{0.036}}
\end{tabular}
   \caption{Visual comparison of our approach against competing methods on the Burst JDD-SR task (best viewed by zooming in). Every output image is accompanied by its LPIPS value.}
\label{fig:burst_images_supp1}
\end{figure*}

\begin{figure*}[!h]
\centering
\setlength\extrarowheight{-10pt}
\begin{tabular}{@{\hskip 0.05cm} c @{\hskip 0.05cm} c @{\hskip 0.05cm} c @{\hskip 0.05cm} c @{\hskip 0.05cm} c @{\hskip 0.05cm}}
  \multirow{3}{*}[1.845cm]{\includegraphics[width=.293\linewidth]{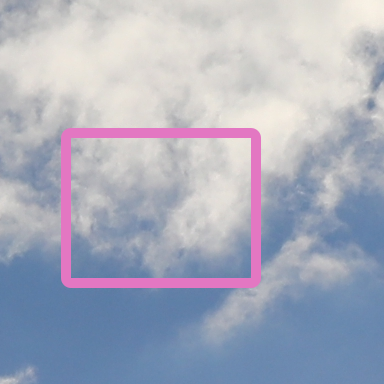}} &
 \includegraphics[width=.170\linewidth]{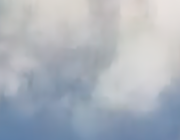} & 
 \includegraphics[width=.170\linewidth]{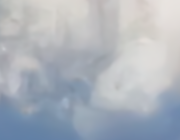} & 
 \includegraphics[width=.170\linewidth]{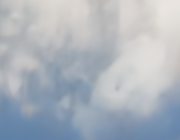} &
 \includegraphics[width=.170\linewidth]{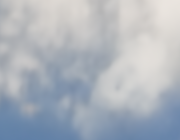} \\
 & \scriptsize{DBSR: 0.280} & \scriptsize{DeepRep: 0.270} & \scriptsize{BSRT-Small: 0.225} & \scriptsize{BSRT-Large: 0.258} \\
 
 & \includegraphics[width=.170\linewidth]{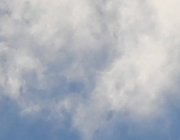} & 
 \includegraphics[width=.170\linewidth]{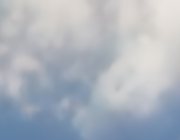} & 
 \includegraphics[width=.170\linewidth]{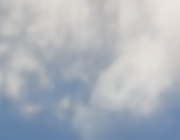} &
 \includegraphics[width=.170\linewidth]{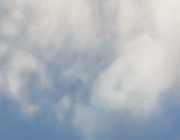} \\
  & \scriptsize{Target} & \scriptsize{BIPNet: 0.258} & \scriptsize{EBSR: 0.252} & \scriptsize{Ours: \textbf{0.131}}
\end{tabular}
   \caption{Visual comparison of our approach against competing methods on the Burst JDD-SR task (best viewed by zooming in). Every output image is accompanied by its LPIPS value.}
\label{fig:burst_images_supp2}
\end{figure*}

\begin{figure*}[!h]
\centering
\setlength\extrarowheight{-10pt}
\begin{tabular}{@{\hskip 0.05cm} c @{\hskip 0.05cm} c @{\hskip 0.05cm} c @{\hskip 0.05cm} c @{\hskip 0.05cm} c @{\hskip 0.05cm}}
  \multirow{3}{*}[1.845cm]{\includegraphics[width=.293\linewidth]{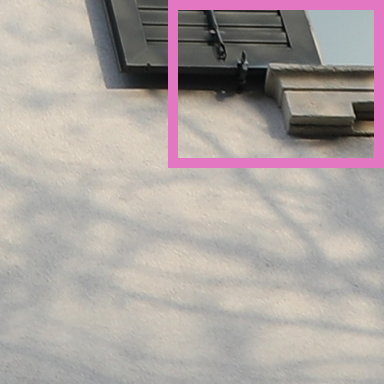}} &
 \includegraphics[width=.170\linewidth]{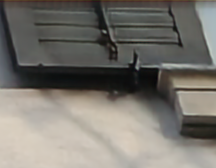} & 
 \includegraphics[width=.170\linewidth]{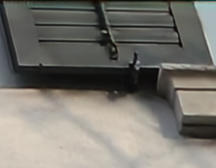} & 
 \includegraphics[width=.170\linewidth]{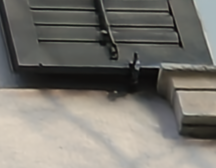} &
 \includegraphics[width=.170\linewidth]{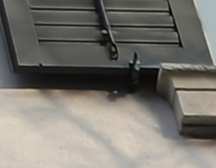} \\
 & \scriptsize{DBSR: 0.160} & \scriptsize{DeepRep: 0.108} & \scriptsize{BSRT-Small: 0.081} & \scriptsize{BSRT-Large: 0.068} \\
 
 & \includegraphics[width=.170\linewidth]{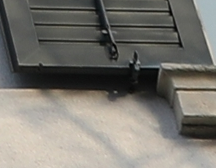} & 
 \includegraphics[width=.170\linewidth]{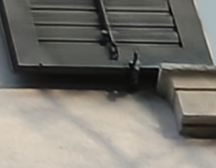} & 
 \includegraphics[width=.170\linewidth]{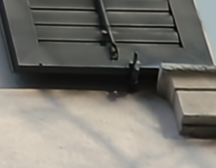} &
 \includegraphics[width=.170\linewidth]{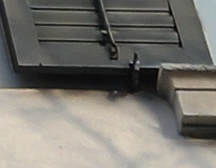} \\
  & \scriptsize{Target} & \scriptsize{BIPNet: 0.089} & \scriptsize{EBSR: 0.083} & \scriptsize{Ours: \textbf{0.042}}
\end{tabular}
   \caption{Visual comparison of our approach against competing methods on the Burst JDD-SR task (best viewed by zooming in). Every output image is accompanied by its LPIPS value.}
\label{fig:burst_images}
\end{figure*}

\begin{figure*}[t!]
\centering
\setlength\extrarowheight{-10pt}
\begin{tabular}{@{\hskip 0.05cm} c @{\hskip 0.05cm} c @{\hskip 0.05cm} c @{\hskip 0.05cm} c @{\hskip 0.05cm} c @{\hskip 0.05cm}}
  \multirow{3}{*}[1.845cm]{\includegraphics[width=.293\linewidth]{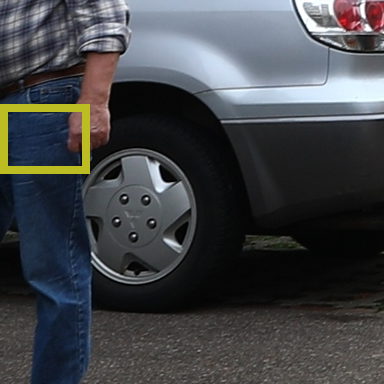}} &
 \includegraphics[width=.170\linewidth]{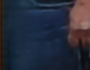} & 
 \includegraphics[width=.170\linewidth]{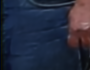} & 
 \includegraphics[width=.170\linewidth]{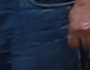} &
 \includegraphics[width=.170\linewidth]{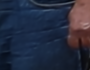} \\
 & \scriptsize{DBSR: 0.165} & \scriptsize{DeepRep: 0.079} & \scriptsize{BSRT-Small: 0.036} & \scriptsize{BSRT-Large: 0.036} \\
 
 & \includegraphics[width=.170\linewidth]{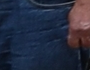} & 
 \includegraphics[width=.170\linewidth]{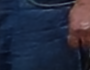} & 
 \includegraphics[width=.170\linewidth]{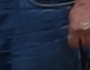} &
 \includegraphics[width=.170\linewidth]{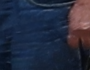} \\
  & \scriptsize{Target} & \scriptsize{BIPNet: 0.043} & \scriptsize{EBSR: 0.035} & \scriptsize{Ours: \textbf{0.030}}
\end{tabular}
   \caption{Visual comparison of our approach against competing methods on the Burst JDD-SR task (best viewed by zooming in). Every output image is accompanied by its LPIPS value.}
   
\label{fig:burst_images}
\end{figure*}

\begin{figure*}[t]
\centering
\setlength\extrarowheight{-10pt}
\begin{tabular}{@{\hskip 0.05cm} c @{\hskip 0.05cm} c @{\hskip 0.05cm} c @{\hskip 0.05cm} c @{\hskip 0.05cm} c @{\hskip 0.05cm} c @{\hskip 0.05cm}}
  \multirow{3}{*}[1.73cm]{\includegraphics[width=.178\linewidth]{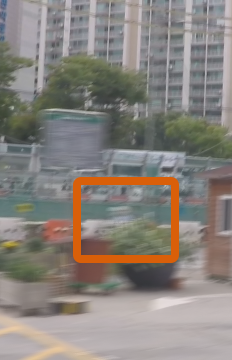}} &
 \includegraphics[width=.159\linewidth]{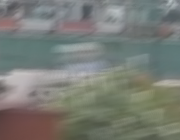} & 
 \includegraphics[width=.159\linewidth]{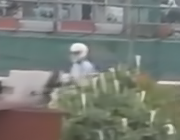} & 
 \includegraphics[width=.159\linewidth]{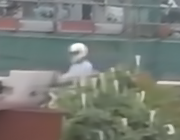} &
 \includegraphics[width=.159\linewidth]{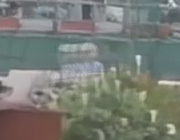} &
 \includegraphics[width=.159\linewidth]{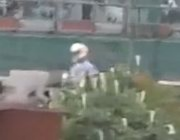}\\
 & \scriptsize{Input: 0.477} & \scriptsize{HINet: 0.128} & \scriptsize{Restormer: 0.135} & \scriptsize{DeblurGANv2: 0.195} & \scriptsize{InDI: 0.111} \\
 
 & \includegraphics[width=.159\linewidth]{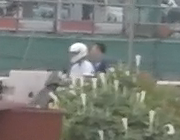} & 
 \includegraphics[width=.159\linewidth]{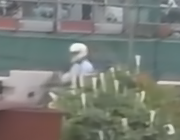} & 
 \includegraphics[width=.159\linewidth]{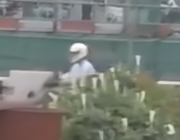} &
 \includegraphics[width=.159\linewidth]{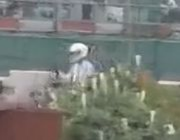} &
 \includegraphics[width=.159\linewidth]{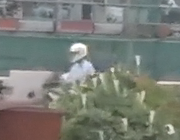} \\
  & \scriptsize{Target} & \scriptsize{NAFNet: 0.112} & \scriptsize{FFTFormer: 0.094} & \scriptsize{DvSR: 0.106} & \scriptsize{Ours: \textbf{0.092}}
\end{tabular}
   \caption{Visual comparison of our approach against competing methods on the GoPro test set for the task of dynamic scene deblurring (best viewed by zooming in). Every output image is accompanied by its LPIPS value.}
\label{fig:gopro_deblur_supp_1}
\end{figure*}

\begin{figure*}[t]
\centering
\setlength\extrarowheight{-10pt}
\begin{tabular}{@{\hskip 0.05cm} c @{\hskip 0.05cm} c @{\hskip 0.05cm} c @{\hskip 0.05cm} c @{\hskip 0.05cm} c @{\hskip 0.05cm} c @{\hskip 0.05cm}}
  \multirow{3}{*}[1.7cm]{\includegraphics[width=.242\linewidth]{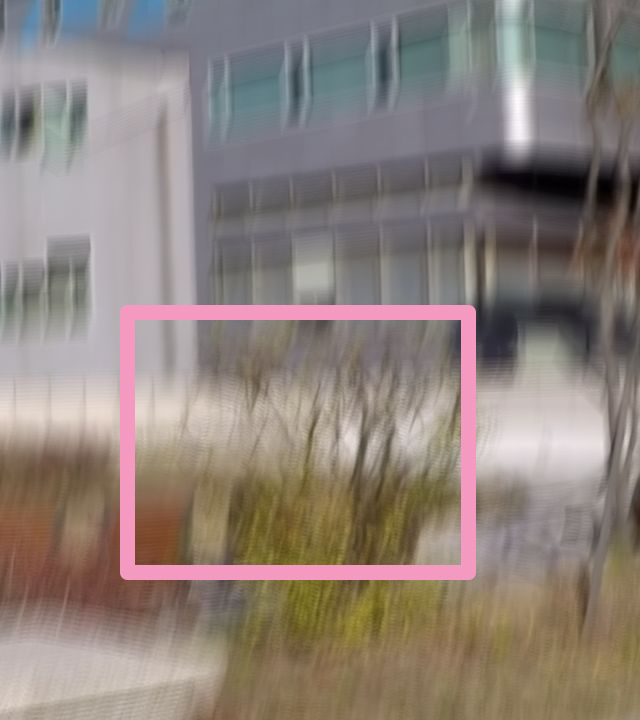}} &
 \includegraphics[width=.159\linewidth]{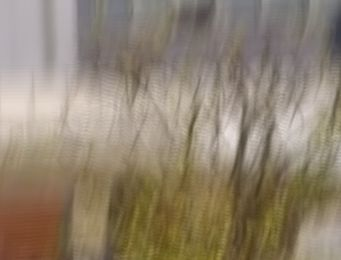} & 
 \includegraphics[width=.159\linewidth]{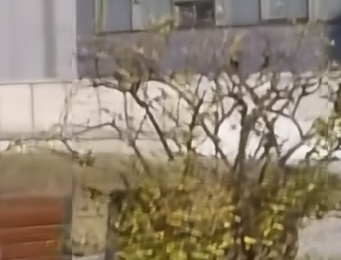} & 
 \includegraphics[width=.159\linewidth]{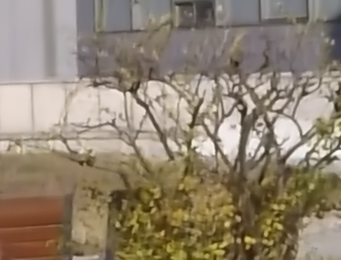} &
 \includegraphics[width=.159\linewidth]{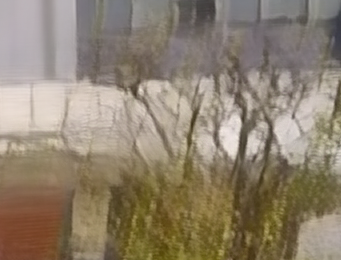} &
 \includegraphics[width=.159\linewidth]{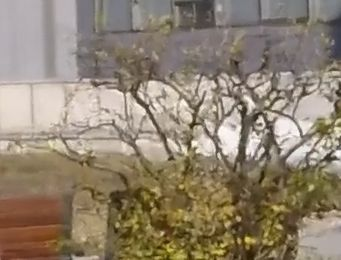}\\
 & \scriptsize{Input: 0.547} & \scriptsize{HINet: 0.171} & \scriptsize{Restormer: 0.137} & \scriptsize{DeblurGANv2: 0.284} & \scriptsize{icDPM: 0.121} \\
 
 & \includegraphics[width=.159\linewidth]{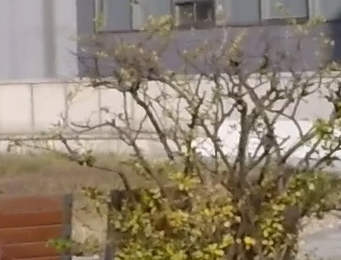} & 
   \includegraphics[width=.159\linewidth]{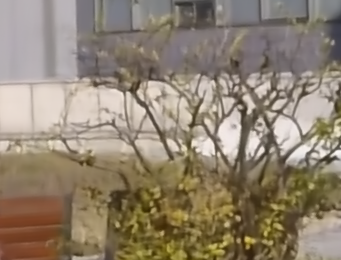} & 
   \includegraphics[width=.159\linewidth]{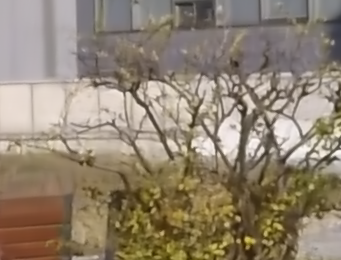} &
   \includegraphics[width=.159\linewidth]{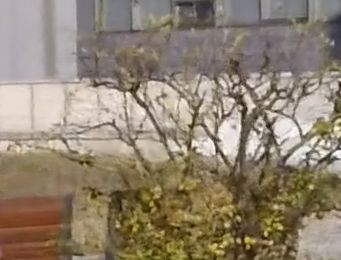} &
   \includegraphics[width=.159\linewidth]{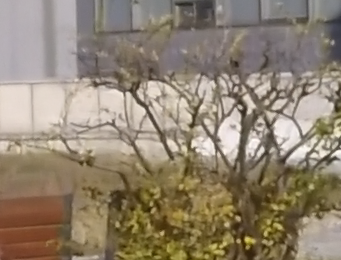} \\
  & \scriptsize{Target} & \scriptsize{NAFNet: 0.133} & \scriptsize{FFTFormer: 0.122} & \scriptsize{DvSR: 0.127} & \scriptsize{Ours: \textbf{0.101}}
\end{tabular}
   \caption{Visual comparison of our approach against competing methods on the GoPro test set for the task of dynamic scene deblurring (best viewed by zooming in). Every output image is accompanied by its LPIPS value.}
\label{fig:GoPro3_deblur}
\end{figure*}

\begin{figure*}[t]
\centering
\setlength\extrarowheight{-10pt}
\begin{tabular}{@{\hskip 0.05cm} c @{\hskip 0.05cm} c @{\hskip 0.05cm} c @{\hskip 0.05cm} c @{\hskip 0.05cm} c @{\hskip 0.05cm} c @{\hskip 0.05cm}}
  \multirow{3}{*}[2.24cm]{\includegraphics[width=.311\linewidth]{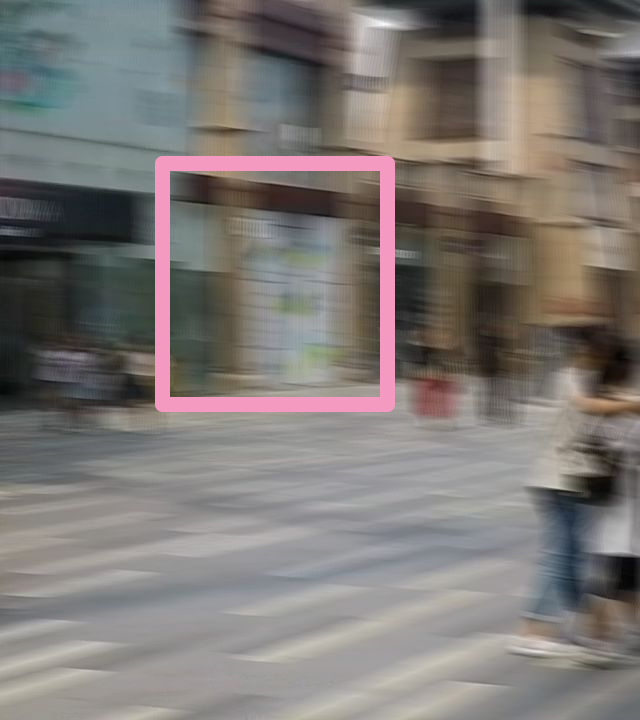}} &
 \includegraphics[width=.15\linewidth]{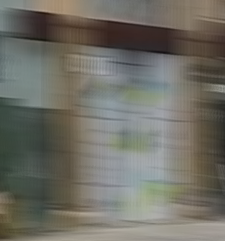} & 
 \includegraphics[width=.15\linewidth]{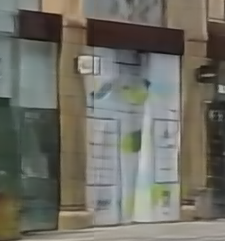} & 
 \includegraphics[width=.15\linewidth]{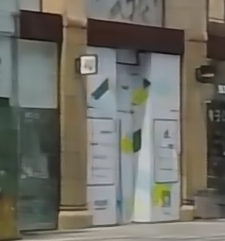} &
 \includegraphics[width=.15\linewidth]{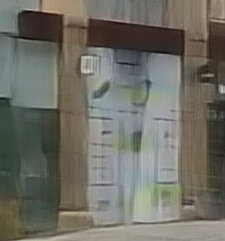} &
 \includegraphics[width=.15\linewidth]{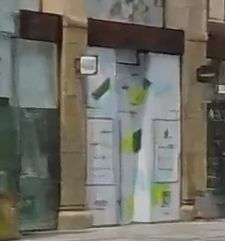}\\
 & \scriptsize{Input: 0.533} & \scriptsize{HINet: 0.157} & \scriptsize{Restormer: 0.095} & \scriptsize{DeblurGANv2: 0.213} & \scriptsize{icDPM: 0.112} \\
 
 & \includegraphics[width=.15\linewidth]{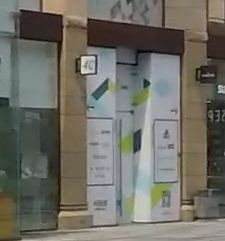} & 
   \includegraphics[width=.15\linewidth]{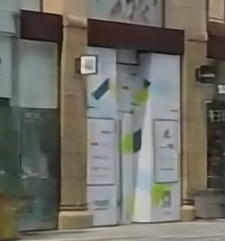} & 
   \includegraphics[width=.15\linewidth]{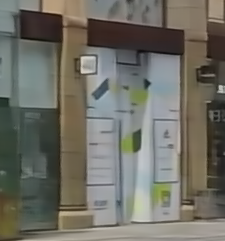} &
   \includegraphics[width=.15\linewidth]{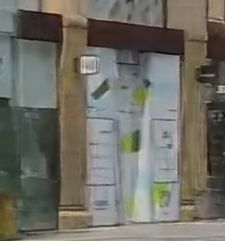} &
   \includegraphics[width=.15\linewidth]{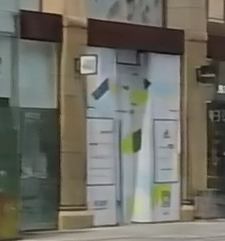} \\
  & \scriptsize{Target} & \scriptsize{NAFNet: \textbf{0.086}} & \scriptsize{FFTFormer: 0.105} & \scriptsize{DvSR: 0.134} & \scriptsize{Ours: 0.097}
\end{tabular}
   \caption{Visual comparison of our approach against competing methods on the HIDE test set for the task of dynamic scene deblurring (best viewed by zooming in). Every output image is accompanied by its LPIPS value.}
\label{fig:HIDE2_deblur}
\end{figure*}

\begin{figure*}[t]
\centering
\setlength\extrarowheight{-10pt}
\begin{tabular}{@{\hskip 0.05cm} c @{\hskip 0.05cm} c @{\hskip 0.05cm} c @{\hskip 0.05cm} c @{\hskip 0.05cm} c @{\hskip 0.05cm} c @{\hskip 0.05cm}}
  \multirow{3}{*}[1.85cm]{\includegraphics[width=.261\linewidth]{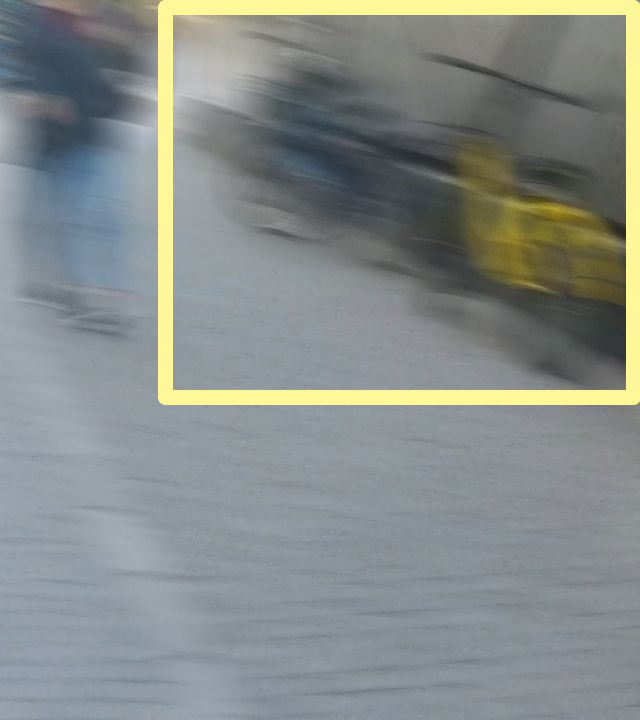}} &
 \includegraphics[width=.159\linewidth]{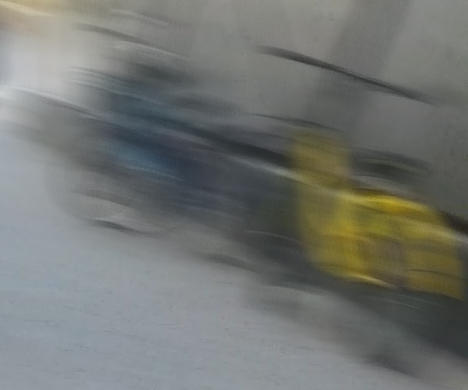} & 
 \includegraphics[width=.159\linewidth]{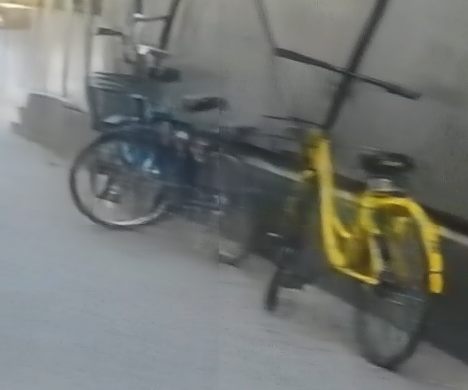} & 
 \includegraphics[width=.159\linewidth]{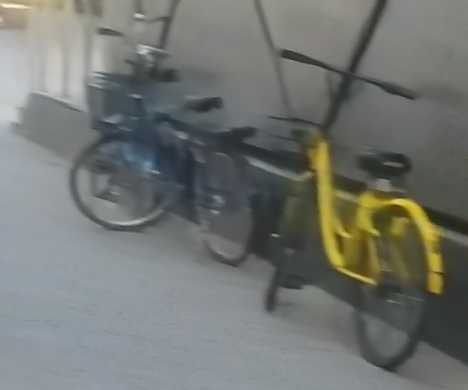} &
 \includegraphics[width=.159\linewidth]{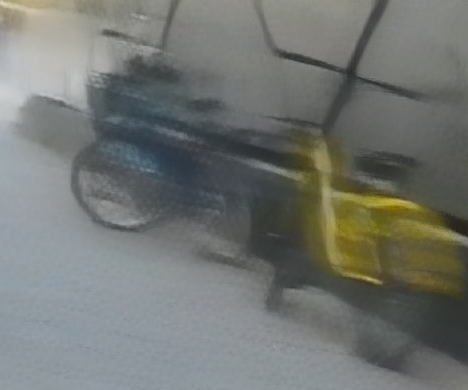} &
 \includegraphics[width=.159\linewidth]{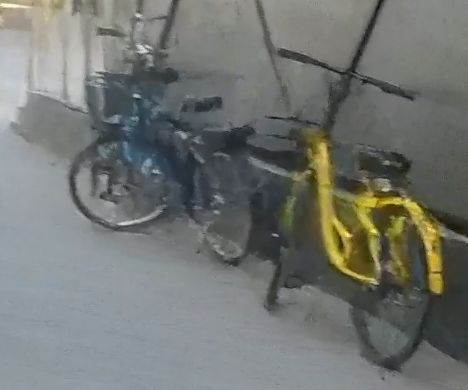}\\
 & \scriptsize{Input: 0.402} & \scriptsize{HINet: 0.242} & \scriptsize{Restormer: 0.248} & \scriptsize{DeblurGANv2: 0.303} & \scriptsize{icDPM: 0.160} \\
 
 & \includegraphics[width=.159\linewidth]{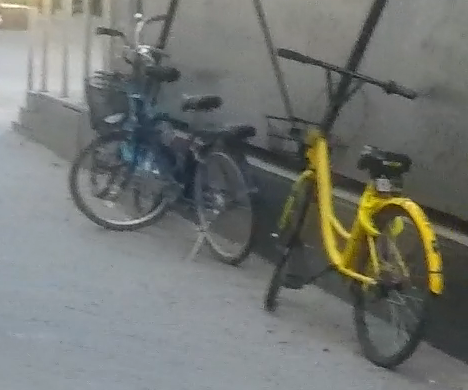} & 
   \includegraphics[width=.159\linewidth]{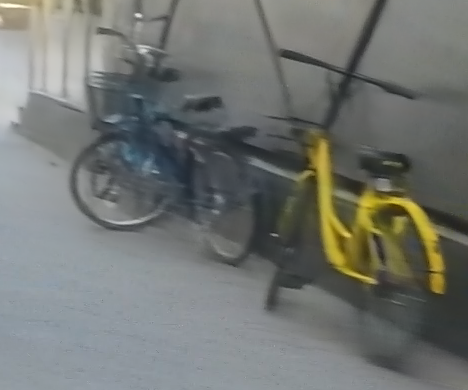} & 
   \includegraphics[width=.159\linewidth]{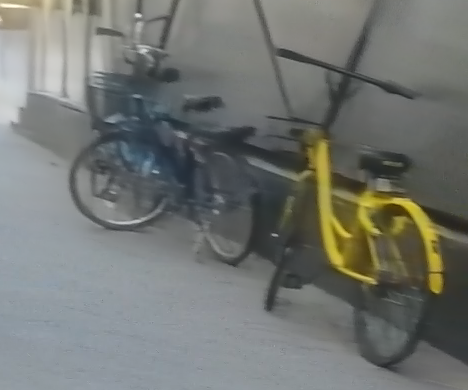} &
   \includegraphics[width=.159\linewidth]{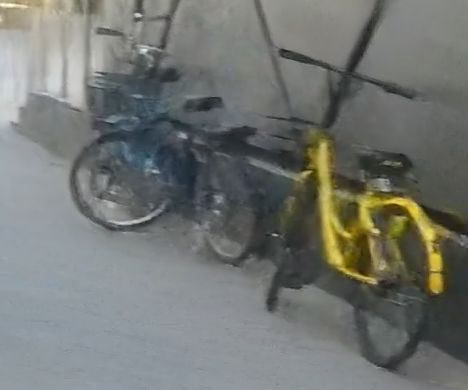} &
   \includegraphics[width=.159\linewidth]{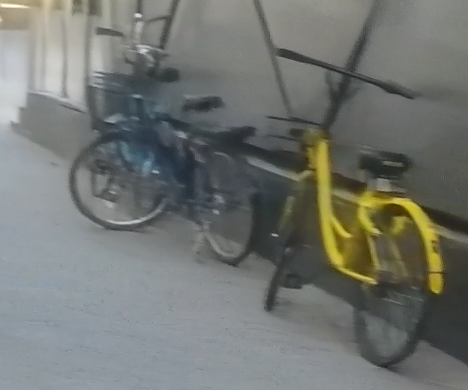} \\
  & \scriptsize{Target} & \scriptsize{NAFNet: 0.207} & \scriptsize{FFTFormer: 0.182} & \scriptsize{DvSR: 0.192} & \scriptsize{Ours: \textbf{0.150}}
\end{tabular}
   \caption{Visual comparison of our approach against competing methods on the HIDE test set for the task of dynamic scene deblurring (best viewed by zooming in). Every output image is accompanied by its LPIPS value.}
\label{fig:HIDE3_deblur}
\end{figure*}

\begin{figure*}[t]
\centering
\setlength\extrarowheight{-10pt}
\begin{tabular}{@{\hskip 0.05cm} c @{\hskip 0.05cm} c @{\hskip 0.05cm} c @{\hskip 0.05cm} c @{\hskip 0.05cm} c @{\hskip 0.05cm} c @{\hskip 0.05cm}}
  \multirow{3}{*}[1.73cm]{\includegraphics[width=.246\linewidth]{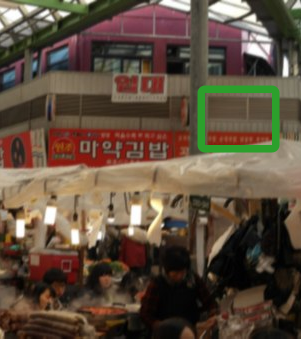}} &
 \includegraphics[width=.159\linewidth]{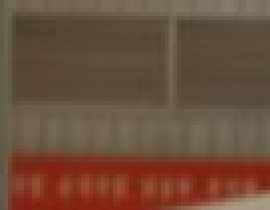} & 
 \includegraphics[width=.159\linewidth]{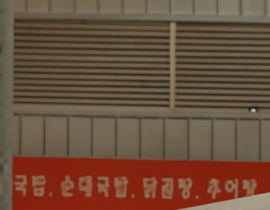} & 
 \includegraphics[width=.159\linewidth]{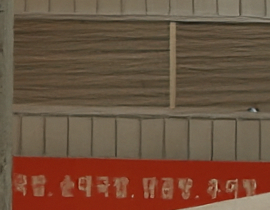} &
 \includegraphics[width=.159\linewidth]{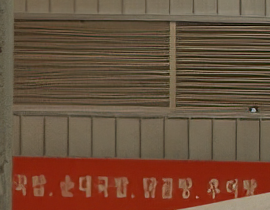} &
 \includegraphics[width=.159\linewidth]{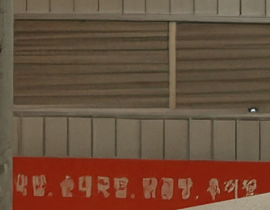}\\
 & \scriptsize{Input: 0.630} & \scriptsize{SwinIR: 0.056} & \scriptsize{HCFlow: 0.208} & \scriptsize{ESRGAN: 0.194} & \scriptsize{InDI: 0.245} \\
 
 & \includegraphics[width=.159\linewidth]{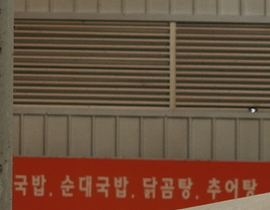} & 
 \includegraphics[width=.159\linewidth]{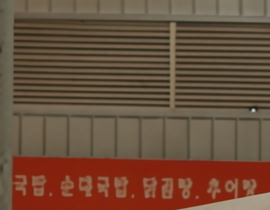} & 
 \includegraphics[width=.159\linewidth]{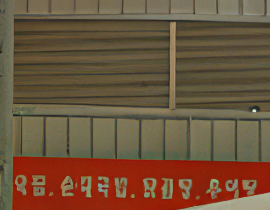} &
 \includegraphics[width=.159\linewidth]{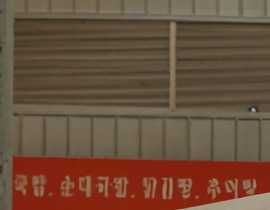} &
 \includegraphics[width=.159\linewidth]{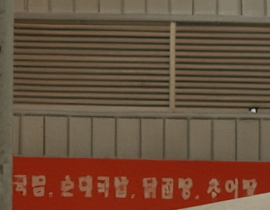} \\
  & \scriptsize{Target} & \scriptsize{HAT: 0.045} & \scriptsize{LDM: 0.273} & \scriptsize{SRDiff: 0.212} & \scriptsize{Ours: \textbf{0.048}}
\end{tabular}
   \caption{Visual comparison of our approach against competing methods on the DIV2K validation set for the task of $4\times$ SISR (best viewed by zooming in). Every output image is accompanied by its LPIPS value.}
\label{fig:div2k_images_supp_1}
\end{figure*}

\begin{figure*}[!t]
\centering
\setlength\extrarowheight{-10pt}
\begin{tabular}{@{\hskip 0.05cm} c @{\hskip 0.05cm} c @{\hskip 0.05cm} c @{\hskip 0.05cm} c @{\hskip 0.05cm} c @{\hskip 0.05cm} c @{\hskip 0.05cm}}
  \multirow{3}{*}[1.73cm]{\includegraphics[width=.246\linewidth]{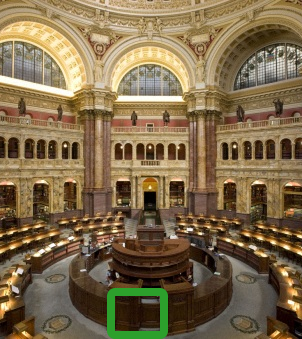}} &
 \includegraphics[width=.159\linewidth]{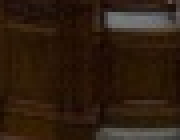} & 
 \includegraphics[width=.159\linewidth]{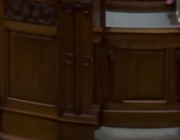} & 
 \includegraphics[width=.159\linewidth]{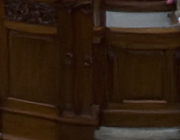} &
 \includegraphics[width=.159\linewidth]{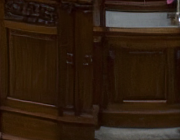} &
 \includegraphics[width=.159\linewidth]{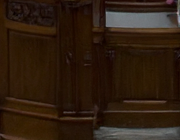}\\
 & \scriptsize{Input: 0.244} & \scriptsize{SwinIR: 0.102} & \scriptsize{HCFlow: 0.075} & \scriptsize{ESRGAN: \textbf{0.072}} & \scriptsize{InDI: 0.074} \\
 
 & \includegraphics[width=.159\linewidth]{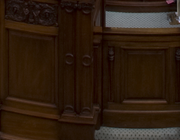} & 
 \includegraphics[width=.159\linewidth]{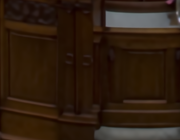} & 
 \includegraphics[width=.159\linewidth]{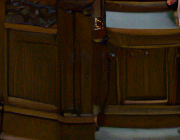} &
 \includegraphics[width=.159\linewidth]{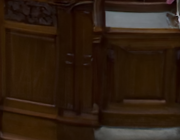} &
 \includegraphics[width=.159\linewidth]{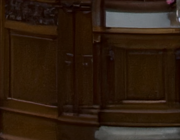} \\
  & \scriptsize{Target} & \scriptsize{HAT: 0.116} & \scriptsize{LDM: 0.190} & \scriptsize{SRDiff: 0.090} & \scriptsize{Ours: \textbf{0.072}}
\end{tabular}
   \caption{Visual comparison of our approach against competing methods on the DIV2K validation set for the task of $4\times$ SISR (best viewed by zooming in). Every output image is accompanied by its LPIPS value.}
\label{fig:div2k_images_supp_2}
\end{figure*}

\begin{figure*}[!t]
\centering
\setlength\extrarowheight{-10pt}
\begin{tabular}{@{\hskip 0.05cm} c @{\hskip 0.05cm} c @{\hskip 0.05cm} c @{\hskip 0.05cm} c @{\hskip 0.05cm} c @{\hskip 0.05cm} c @{\hskip 0.05cm}}
  \multirow{3}{*}[1.73cm]{\includegraphics[width=.246\linewidth]{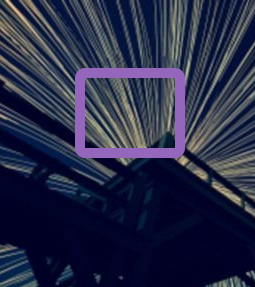}} &
 \includegraphics[width=.159\linewidth]{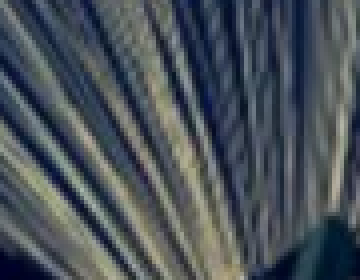} & 
 \includegraphics[width=.159\linewidth]{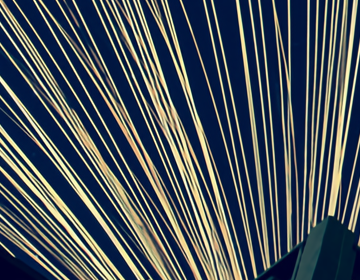} & 
 \includegraphics[width=.159\linewidth]{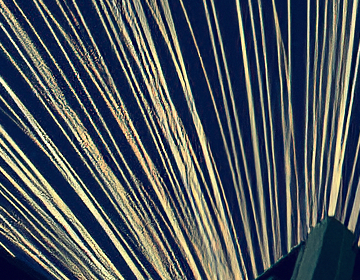} &
 \includegraphics[width=.159\linewidth]{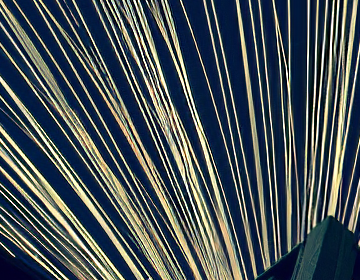} &
 \includegraphics[width=.159\linewidth]{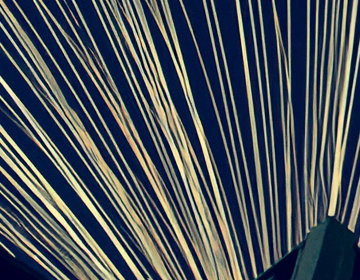}\\
 & \scriptsize{Input: 0.394} & \scriptsize{SwinIR: 0.114} & \scriptsize{HCFlow: 0.046} & \scriptsize{ESRGAN: 0.044} & \scriptsize{InDI: 0.057} \\
 
 & \includegraphics[width=.159\linewidth]{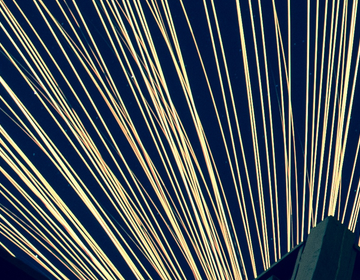} & 
 \includegraphics[width=.159\linewidth]{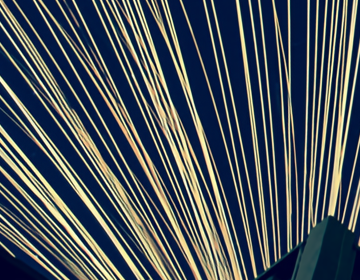} & 
 \includegraphics[width=.159\linewidth]{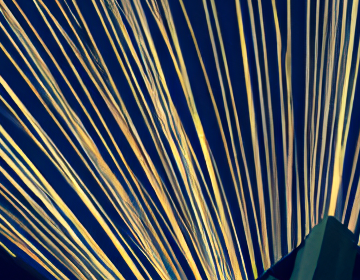} &
 \includegraphics[width=.159\linewidth]{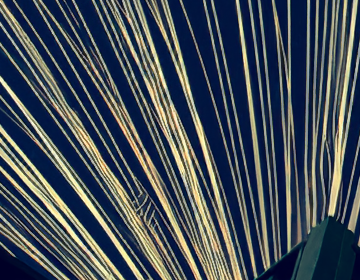} &
 \includegraphics[width=.159\linewidth]{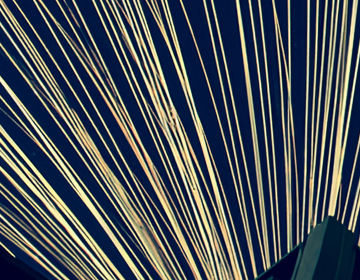} \\
  & \scriptsize{Target} & \scriptsize{HAT: \textbf{0.022}} & \scriptsize{LDM: 0.141} & \scriptsize{SRDiff: 0.061} & \scriptsize{Ours: \textbf{0.022}}
\end{tabular}
   \caption{Visual comparison of our approach against competing methods on the DIV2K validation set for the task of $4\times$ SISR (best viewed by zooming in). Every output image is accompanied by its LPIPS value.}
\label{fig:div2k_images_supp_3}
\end{figure*}

\begin{figure*}[t]
\centering
\setlength\extrarowheight{-10pt}
\begin{tabular}{@{\hskip 0.05cm} c @{\hskip 0.05cm} c @{\hskip 0.05cm} c @{\hskip 0.05cm} c @{\hskip 0.05cm} c @{\hskip 0.05cm} c @{\hskip 0.05cm}}
  \multirow{3}{*}[1.62cm]{\includegraphics[width=.261\linewidth]{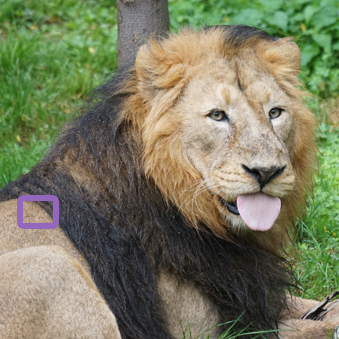}} &
 \includegraphics[width=.155\linewidth]{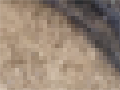} & 
  \includegraphics[width=.155\linewidth]{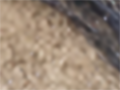} & 
 \includegraphics[width=.155\linewidth]{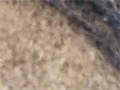} &
 \includegraphics[width=.155\linewidth]{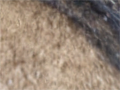} &
 \includegraphics[width=.155\linewidth]{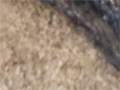}\\
 & \scriptsize{Input: 0.335} & \scriptsize{SwinIR: 0.230} & \scriptsize{HCFlow: \textbf{0.114}} & \scriptsize{ESRGAN: 0.127} & \scriptsize{InDI: 0.194} \\
 
 & \includegraphics[width=.155\linewidth]{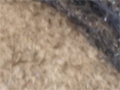} & 
\includegraphics[width=.155\linewidth]{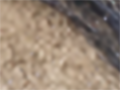} & 
\includegraphics[width=.155\linewidth]{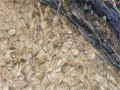} &
  \includegraphics[width=.155\linewidth]{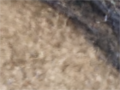} &
 \includegraphics[width=.155\linewidth]{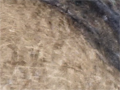} \\
  & \scriptsize{Target} & \scriptsize{HAT: 0.219} & \scriptsize{LDM: 0.269} & \scriptsize{SRDiff: 0.119} & \scriptsize{Ours: 0.140}
\end{tabular}
   \caption{Visual comparison of our approach against competing methods on the DIV2K validation set for the task of $4\times$ super-resolution (best viewed by zooming in). Every output image is accompanied by its LPIPS value.}
\label{fig:div2k_images_supp1}
\end{figure*}

\end{document}